\documentclass[10pt,twocolumn,letterpaper]{article}
\usepackage{cvpr}
\usepackage[pdftex]{graphicx}
\usepackage{amsmath}
\usepackage{amssymb}
\usepackage{booktabs}
\usepackage{multirow}
\usepackage{cuted}
\usepackage{xspace}
\usepackage{wasysym}
\usepackage{manfnt}
\usepackage[accsupp]{axessibility}
\usepackage[normalem]{ulem}
\usepackage[dvipsnames]{xcolor}
\usepackage[pagebackref,breaklinks,colorlinks]{hyperref}
\usepackage[export]{adjustbox}

\usepackage{amssymb}% http://ctan.org/pkg/amssymb
\usepackage{pifont}% http://ctan.org/pkg/pifont
\newcommand{\cmark}{\ding{51}}%
\newcommand{\xmark}{\ding{55}}%

\usepackage[capitalize]{cleveref}
\crefname{section}{Sec.}{Secs.}
\Crefname{section}{Section}{Sections}
\Crefname{table}{Table}{Tables}
\crefname{table}{Tab.}{Tabs.}
\Crefname{equation}{Equation}{Equations}
\crefname{equation}{eq.}{eqs.}

\makeatletter
\renewcommand{\paragraph}{%
  \@startsection{paragraph}{4}%
  {\z@}{0.25em}{-1em}%
  {\normalfont\normalsize\bfseries}%
}
\makeatother

\def\cvprPaperID{8155}
\def\confName{CVPR}
\def\confYear{2023}

\makeatletter
\DeclareRobustCommand\onedot{\futurelet\@let@token\@onedot}
\def\@onedot{\ifx\@let@token.\else.\null\fi\xspace}

\def\etal{\emph{et al}\onedot}
\makeatother

\begin{document}
\title{DynamicStereo: Consistent Dynamic Depth from Stereo Videos}
\author{
Nikita Karaev$^{1,2}$~~~~~~~Ignacio Rocco$^{1}$~~~~~~~Benjamin Graham$^{1}$~~~~~~~Natalia Neverova$^{1}$\\
Andrea Vedaldi$^{1}$~~~~~~~Christian Rupprecht$^{2}$
\vspace{1em}
\\
$^1$ Meta AI~~~~~~~~~~$^2$ Visual Geometry Group, University of Oxford
\vspace{-4ex}
}

\maketitle
\begin{strip}
\centering
\includegraphics[width=2.1\columnwidth]{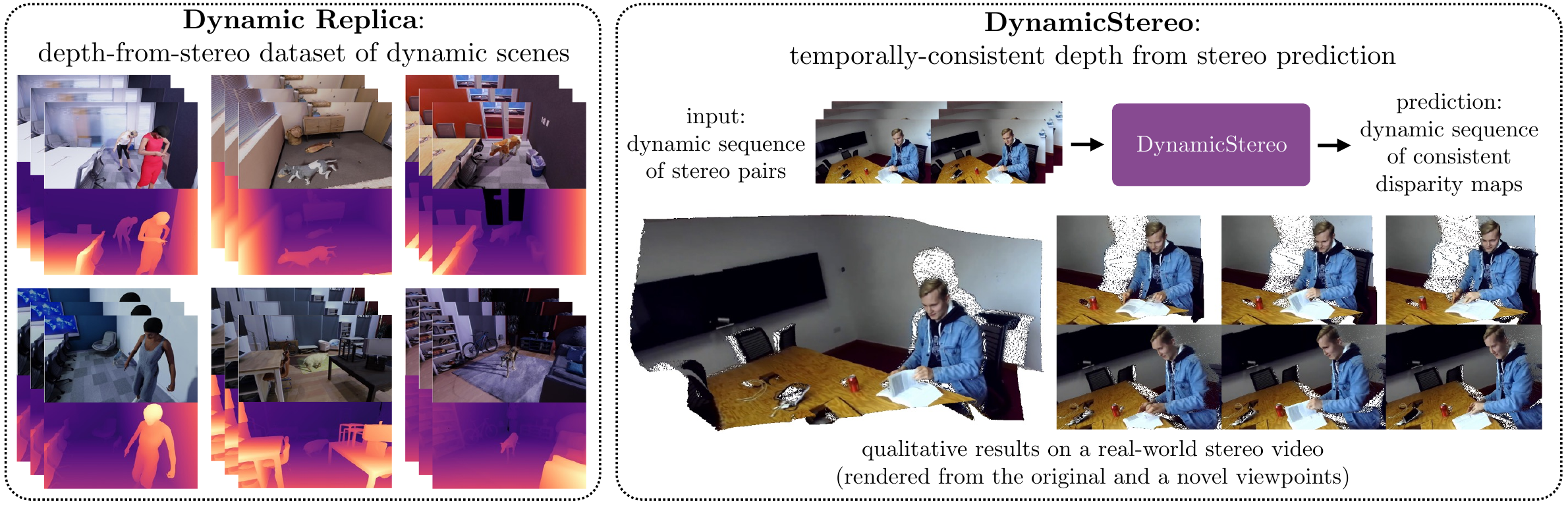}
\captionof{figure}{\textbf{Dynamic Replica and DynamicStereo.} In this work, we (a) introduce a new synthetic stereo video dataset (Dynamic Replica) to train and benchmark temporally consistent disparity estimators for dynamic scenes with people and animals, and (b) propose a method (DynamicStereo) exploiting recent advances in transformer architectures that performs efficient stereo matching at the level of videos.} 
\label{f:splash}
\end{strip}

\thispagestyle{empty} \pagestyle{empty}

\begin{center}
  \large\textbf{Abstract}
\end{center}

\textit{
We consider the problem of reconstructing a dynamic scene observed from a stereo camera.
Most existing methods for depth from stereo treat different stereo frames independently, leading to temporally inconsistent depth predictions.
Temporal consistency is especially important for immersive AR or VR scenarios, where flickering greatly diminishes the user experience. 
We propose DynamicStereo, a novel transformer-based architecture to estimate disparity for stereo videos. 
The network learns to pool information from neighboring frames to improve the temporal consistency of its predictions.
Our architecture is designed to process stereo videos efficiently through divided attention layers.
We also introduce Dynamic Replica, a new benchmark dataset containing synthetic videos of people and animals in scanned environments, which provides complementary training and evaluation data for dynamic stereo closer to real applications than existing datasets. Training with this dataset further improves the quality of predictions of our proposed DynamicStereo as well as prior methods. 
Finally, it acts as a benchmark for consistent stereo methods.
Project page: \href{https://dynamic-stereo.github.io/}{https://dynamic-stereo.github.io/}}

\section{Introduction}%
\label{sec:intro}

Estimating depth from stereo is a fundamental computer vision problem,
with applications in 3D reconstruction, robot navigation, and human motion capture, among others.
With the advent of consumer devices featuring multiple cameras, such as AR glasses and smartphones, stereo can simplify the 3D reconstruction of everyday scenes, extracting them as content to be experienced in virtual or mixed reality, or for mixed reality pass-through\@.

Depth from stereo takes as input two images capturing the same scene from different viewpoints.
It then finds pairs of matching points, a problem known as~\emph{disparity estimation}.
Since the two cameras are calibrated, the matched points can be projected into 3D using triangulation.
While this process is robust, it is suboptimal when applied to video data, as it can only reconstruct stereo frames individually, ignoring the fact that the observations infer properties of the \emph{same underlying objects} over time.
Even if the camera moves or the scene deforms non-rigidly, the instantaneous 3D reconstructions are highly correlated and disregarding this fact can result in inconsistencies.

In this paper, we thus consider the problem of \emph{dynamic depth from stereo} to improve the temporal consistency of stereo reconstruction from video data.

Traditional approaches to stereo compute the matching costs between local image patches, aggregating those in an objective function, and optimizing the latter together with a regularization term to infer disparities.
Examples of such approaches include max-flow~\cite{maxflow} and graph-cut~\cite{kolmogorov2001computing}.
More recently, stereo methods have used deep networks learned from a large number of image pairs annotated with ground-truth disparities~\cite{kendall2017end, guo2019group, raft_stereo,CREStereoLi2022practical}.
They usually follow an approach similar to the traditional methods, but using deep CNN features for computing the matching costs, and replacing the per-image optimization by a pre-trained regression deep network, which processes the cost volume and outputs the estimated disparities.

In the video setting, matching quality can potentially be improved by looking for matches across space \emph{and} time.
For instance, points occluded in one camera at a given point in time may be visible from both cameras at other times. 

Transformer architectures have shown that attention can be a powerful and flexible method for pooling information over a range of contexts~\cite{bert, carion2020end, sun2021loftr, dosovitskiy2021image}.
Our \emph{DynamicStereo} model incorporates self- and cross-attention to extract relevant information across space, time and stereo pairs.
Our architecture relies on divided attention~\cite{gberta_2021_ICML} to allow efficient processing of this high-dimensional space.

\begin{figure}
\centering
\includegraphics[width=\columnwidth]{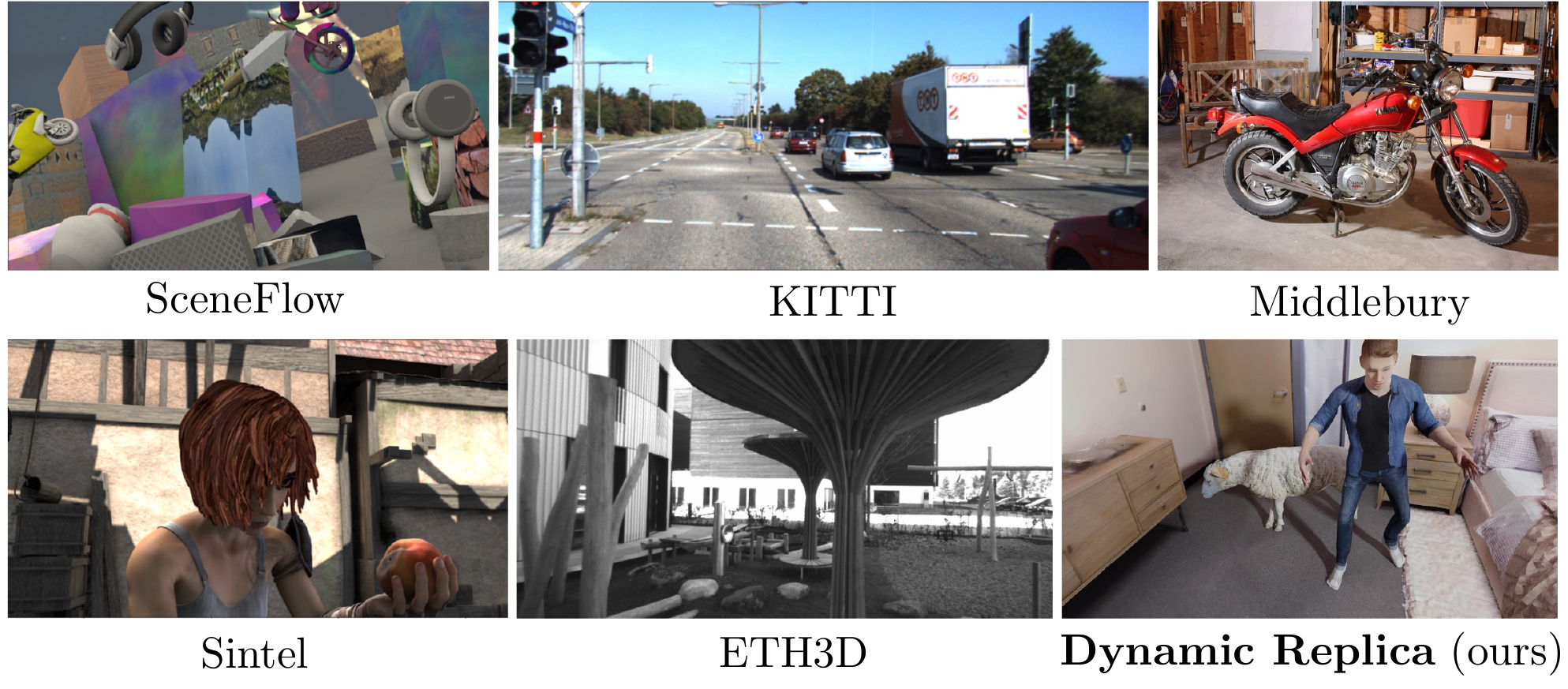}
\caption{\textbf{Example frames from depth-from-stereo datasets.} We visually compare 
current datasets to \emph{Dynamic Replica}, which contains renderings of every-day scenes with people and animals and differs from existing datasets in size, realism, and content.}\label{fig:all_datasets}
\end{figure}

As a learning-based approach, we wish to learn priors from data representative of real-life 3D dynamic reconstruction applications, where videos depict people or animals moving and interacting with objects.
There are several synthetic video datasets~\cite{Mayer_2016, AutoFlow, flyingchairs} commonly used for training stereo and optical flow methods, but they contain  abstract scenes with several layers of moving objects that share little resemblance to real-life.
More realistic stereo datasets also exist~\cite{falling_things, tartanair}, but they either do not contain video sequences or are focused on static scenes.
Given these limitations, as an additional contribution we propose a new synthetic stereo dataset showing moving human and animal characters inside realistic physical spaces from the Replica dataset~\cite{replica19arxiv}.
We call this new dataset \emph{Dynamic Replica} (DR), and we use it for learning dynamic stereo matches.
DR contains 524 videos of virtual humans and animals embedded in realistic digital scans of physical environments (see \cref{table:datasets_statistics} and \cref{fig:all_datasets}).
We show that DR can significantly boost the quality of dynamic stereo methods compared to training them only on existing depth-from-stereo datasets.

To summarise, we make \textbf{three contributions}.
(1) We introduce \emph{DynamicStereo}, a transformer-based architecture that improves dynamic depth from stereo by jointly processing stereo videos.
(2) We release \emph{Dynamic Replica}, a new benchmark dataset for learning and evaluating models for dynamic depth from stereo.
(3) We demonstrate state-of-the-art dynamic stereo results in a variety of benchmarks.

\section{Related work}%
\label{sec:related}

\begin{table*}[t]
  \centering
  \footnotesize
  \begin{tabular}{lcccccc}
      \toprule
      Dataset property & MPI Sintel \cite{Sintel} & KITTI \cite{Kitti} & SceneFlow \cite{Mayer_2016} & Falling Things \cite{falling_things} & TartanAir \cite{wang2020tartanair} & Dynamic Replica (ours) \\
      \midrule
      \#Training frames  & $1\,064$                         & $194$+$200$  & $34\,801$ & $60\,200$ &  $296\,000$ & $145\,200$\\
      \#Test frames      & $564$                & $195$+$200$             &  $4\,248$  & $0$ & 0 &$6\,000 + 18\,000 $ \\
      \#Training sequences & $25$                 & $194$+$200$             &  $2\,256$ & ($330$) & $1037$ & $484$ \\
      Resolution         & $1024\!\times\!436$  & $1242\!\times\!375$  & $960\!\times\!540$ & $960\!\times\!540$ & $640\!\times\!480$ & $1280\!\times\!720$ \\
      \midrule
      Disparity/Depth    & \cmark   & sparse     & \cmark   & \cmark  & \cmark & \cmark \\
      Optical flow       & \cmark   & (sparse)   & \cmark   & \xmark  & \cmark  &  \cmark \\
      Segmentation       & \cmark   & \xmark  & \cmark   & \cmark  & \cmark & \cmark \\
      \midrule 
      Non-rigid objects  &  \cmark  & \xmark   &         (\cmark)    & \xmark & (\xmark)  &\cmark  \\
      Realism         & (\cmark) & \cmark      & \xmark & (\cmark) &  (\cmark)  & (\cmark) \\
      \bottomrule
    \end{tabular}
  \caption{\textbf{Comparison of depth-from-stereo datasets.} \emph{Dynamic Replica} is larger than previous datasets in both resolution and number of frames. A main aspect is that it contains non-rigid objects such as animals and people, which is not available at scale in prior datasets.}
  \label{table:datasets_statistics}
\end{table*}

\paragraph{Depth from stereo.}

Stereo matching is a classic problem in computer vision.
The traditional way of solving it is to compute local matching costs between image patches and then either perform local aggregation~\cite{birchfield1999depth, van2002hierarchical, hirschmuller2002real, zitnick2000cooperative}, or a global optimization based on an energy function featuring data and regularization terms\cite{maxflow,kolmogorov2001computing}.

More recently, deep learning has become the dominant approach to stereo matching.
Zbontar and LeCun~\cite{DBLP:journals/corr/ZbontarL14} proposed to use a CNN to compute the matching cost between image patches. 
Then Mayer~\etal~\cite{Mayer_2016} introduced the first fully learning-based approach to stereo estimation. Inspired by traditional stereo algorithms, the next line of work \cite{yao2018mvsnet, chang2018pyramid, kendall2017end, zhang2019ga, guo2019group, zhang2020domain} applied 3D convolutions in their fully learning-based approaches.  They construct a dense 3D cost volume between the left and the right 2D feature-maps before filtering it with a 3D-CNN. These approaches often fail to generalize to data that they were not trained on.

More recent works focused on improving the computational efficiency of stereo estimation \cite{tankovich2021hitnet, liang2018learning, badki2020bi3d, raft_stereo}.  
Inspired by RAFT \cite{RaftTeed021} which constructs a 4D cost volume between all pairs of pixels for optical flow, RAFT-Stereo~\cite{raft_stereo} restricts this volume to 3D by collapsing it to the epipolar line. Similarly to RAFT, it iteratively updates the prediction at high resolution allowing to maintain global context and recover details. 
% RAFT-Stereo~\cite{raft_stereo} restricts 
These updates are performed by a GRU~\cite{BallasYPC15} that operates at multiple resolutions.

CRE-Stereo~\cite{CREStereoLi2022practical} proposes to gradually increase resolution during iterative updates which simplifies the architecture. However, the lack of information propagation from lower to higher resolution may result in a loss of context. Our architecture also refines disparity in a coarse-to-fine manner, but unlike CRE-Stereo, we fuse low-resolution and high-resolution features together and propagate them to the final prediction. This allows high-resolution layers to use these features and keep track of previous states.

Some recent works incorporate attention~\cite{attention}. 
Stereo Transformer~\cite{STTR} replaces cost volume with dense pixel matching using attention. LoFTR~\cite{sun2021loftr} and SuperGlue \cite{sarlin2020superglue} use combinations of self and cross-attention layers for sparse feature matching.
Inspired by these methods, CRE-Stereo~\cite{CREStereoLi2022practical} uses self and cross-attention to improve convolutional features.
These works focus on disparity estimation for individual frames. As we predict disparity for videos, we apply attention across time, space, and stereo frames.

\paragraph{Dynamic video depth.}

Monocular depth estimators like MiDaS~\cite{ranftl2020towards, midas} attempt to estimate depth from a single image.
Due to the ambiguity of this task, they tend to be far less accurate than methods that use stereo.
Recent works have proposed to extend these methods to use monocular \emph{videos} instead, relying on motion parallax in static areas~\cite{li2019learning} or fusing information extracted from all the frames of a video, in an off-line fashion.

Consistent Video Depth (CVD)~\cite{cvd} assumes that objects move almost rigidly across neighboring frames.
Robust CVD (RCVD)~\cite{rcvd} excludes dynamic objects from the optimization objective.
Dynamic VD (DVD)~\cite{dvd} explicitly models motion of non-rigid objects with a scene-flow network.
All these methods require fine-tuning a monocular depth predictor like MiDAS on a specific video, a process that must be carried out from scratch for each new video.

\paragraph{Dynamic depth from stereo.}
The work of Li~\etal (CODD)~\cite{Holograms} is the closest to ours.
It is based on three different networks: stereo, motion, and fusion, that are trained separately.
During inference, consistency is achieved by extracting information from the memory state that is updated after each iteration.
It is an online approach that assumes no access to future frames and thus does not allow global optimization over the whole sequence.
Our network does not need a memory state and learns consistency from data. It can be applied in both online and offline settings.

\paragraph{Datasets for learning depth from stereo.}

Training data is an important factor for learning-based stereo algorithms.
It is challenging to collect real data for this task because disparity is extremely difficult to annotate.~\cite{wang2019web, Kitti}
Active sensors such as time-of-flight cameras can provide ground truth depth.
KITTI~\cite{Kitti} shows that converting such ground truth depth to pixel-accurate annotations is still challenging for dynamic objects due to the low frame rate and imprecise estimations.
Synthetic datasets~\cite{richter2016playing, richter2017playing, krahenbuhl2018free} can simplify the data collection process.
SceneFlow~\cite{Mayer_2016} is the first large-scale synthetic dataset for disparity and optical flow estimation.
It allowed training fully learning-based methods for these tasks.
SceneFlow is less realistic compared to MPI Sintel~\cite{Sintel}, a much smaller dataset with optical flow and disparity annotations.
Falling Things~\cite{falling_things} is realistic and relatively large but does not contain non-rigid objects. TartanAir\cite{tartanair} is a realistic SLAM dataset with ground-truth stereo information but only few non-rigid objects.

To the best of our knowledge, we are the first to introduce a large-scale semi-realistic synthetic dataset with a focus on non-rigid objects for disparity estimation.

\begin{figure*}
\centering
\includegraphics[width=\textwidth]{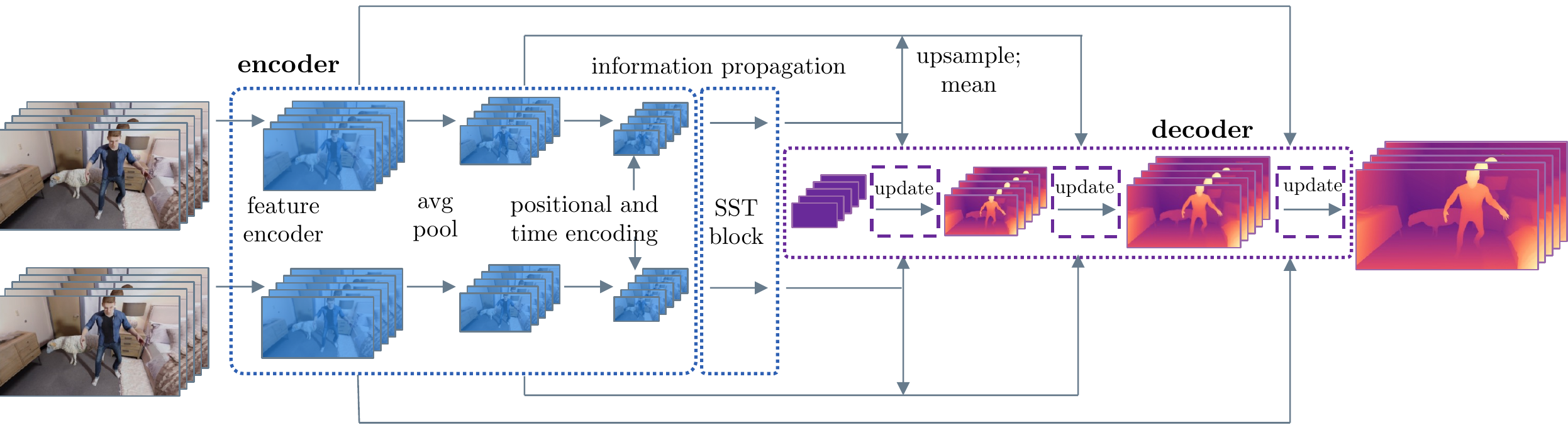}
\caption{\textbf{Proposed architecture of DynamicStereo.} Our method consists of three main components. A convolutional encoder produces feature maps at three scales. As data spans time, space, and stereo views, the SST-Block operating at the smallest resolution, ensures information exchange across all dimensions. Finally, the prediction is generated by a decoder composed of an iterative update function $g$.}\label{fig:architecture}
\end{figure*}

\section{Method}%
\label{sec:overview}

We first formalize the \emph{dynamic depth from stereo} problem.
Given a sequence $S = [(I^L_t, I^R_t)]_{1 \leq t \leq T} \in \mathbb{R}^{2 \times 3 \times H \times W} $ of $T$ rectified stereo frames, the task is to predict a sequence of disparity maps $\hat{D} = [D_t]_{1 \leq t \leq T} \in \mathbb{R}^{H \times W}$ aligned with the left frames $I^L_t$.
While most disparity estimation methods operate on single frames $\hat{D}_t = \Phi(S_t)$, here we learn a model that operates on sequences of length $T$, constructing a function $\hat{D} = \Phi(S)$.
This has the advantage that the model can fuse information along time and thus improve its temporal consistency.

The challenge is to design an architecture that can pass information across such a large volume of data efficiently.
We achieve this via an encoder-decoder design (\cref{fig:architecture}), building on prior work~\cite{unet, flownet, CREStereoLi2022practical}.
The encoder extracts features independently from all the frames and obtains a multi-scale representation of their content.
The decoder then matches progressively more detailed features to recover the disparities from coarse to fine.
Low-resolution matches easily spot large displacements of large image regions, capturing the rough structure of the disparity map, whereas high-resolution matches recover the details.

For efficiency, matches are always carried out along epipolar lines.
The task of exchanging information across space, view and time is delegated to two mechanisms.
First, the encoder terminates in a transformer network that updates the lowest-resolution feature by running attention across these three dimensions in turn.
This is efficient because it is done only for the lowest-resolution features.
Second, the decoder further propagates information as it matches features to recover the disparities. The decoder itself consists of update blocks (\cref{fig:update_block}a) that use both space and time information to gradually refine the disparity.

\subsection{Encoder}

The encoder starts by \textbf{extracting features} from every frame $I^v_t$, $v \in \{L,R\}$, $t \in \{1 ,\dots,T\}$ of the stereo video independently by applying the same CNN $F$ to them.
As it is typical for CNN backbones, the resolution of the output feature map is lower than the resolution of the input image.
Features are extracted with minimum stride $k=4$ and are further down-sampled with average pooling to obtain features at $1/8$ and $1/16$ of the original resolution.
Overall, this results in the feature maps
$
\phi(I^{v}_t)_k \in \mathbb{R}^{d\times \frac{H}{k} \times \frac{W}{k}}
$,
$
k \in \{4,8,16\}
$.
We will use the symbol $\phi_k \in \mathbb{R}^{T \times 2 \times d \times \frac{H}{k} \times \frac{W}{k}}$ to refer to the combined five-dimensional (temporal, stereo and spatial) feature volume at resolution $k$.

\begin{figure*}
\centering
\includegraphics[width=\textwidth]{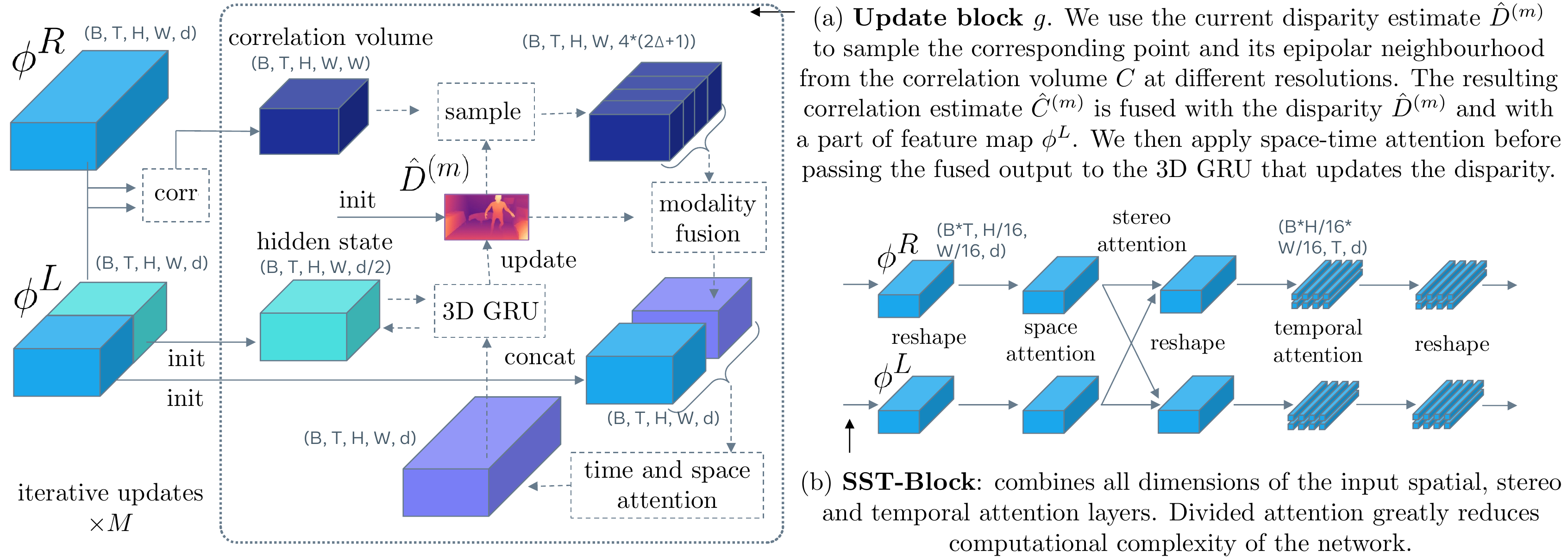}
\caption{\textbf{The proposed architecture of (a) update block $g$ and (b) SST-Block.} 
\label{fig:update_block}}
\end{figure*}

\paragraph{Information Propagation.}
The backbone $F$ processes frames independently, so we require a different mechanism to exchange of information between left and right views $v$ and different timestamps $t$.  We do so by passing the features to a transformer network that uses self and cross-attention. 
Ideally, attention should compare feature maps across views, time, and space.
However, it is computationally demanding to apply attention jointly even with linear~\cite{sun2021loftr} space and stereo attention. We thus rely on divided attention~\cite{gberta_2021_ICML} to attend these three dimensions individually.
We call this a Space-Stereo-Time attention block (SST, \cref{fig:update_block}b) and we repeat it four times.
Since attention remains computationally expensive, we only apply SST to the lowest-resolution feature map $\phi_{16}$ only.

\subsection{Decoder}

The output of the encoder is a multi-resolution feature volume, where the lowest resolution features incorporate information across time, view and space dimensions due to the SST block described above.
The task of the decoder is to convert this feature volume into the final disparity values.

The decoder is based on three ideas:
(1) Disparities are updated from coarse to fine~\cite{CREStereoLi2022practical}, using features of increasing resolution to update earlier estimates.
(2) At each resolution, a feature correlation volume is computed and correspondences are refined iteratively~\cite{RaftTeed021} with reference to this volume.
(3) Similar to the SST attention blocks in the encoder, information is exchanged between the three dimensions (space, view and time) throughout decoding.
We describe next each component in detail.

\paragraph{Iterative correspondence updates.}
Our model produces a sequence of progressively more accurate disparity estimates $\hat{D} = \hat D^{(m)} \in \mathbb{R}^{\frac{H}{k} \times \frac{W}{k}}$, $0 \leq m \leq M$ starting from $\hat D^{(0)} = 0$ and then applying $M$ times the update rule:
\begin{equation}\label{eq:update}
\hat D^{(m+1)} = \hat D^{(m)} + g(\hat D^{(m)}, \phi).
\end{equation}

We apply the rule $\frac{M}{4}$ times to obtain disparities $\hat D^{(m)}$ at the coarser resolution $(\frac{H}{16},\frac{W}{16})$.
We then upsample $\hat D^{(\frac{M}{4})}$ to resolution $(\frac{H}{8},\frac{W}{8})$, apply the update rule $\frac{M}{4}$ more times to obtain $\hat D^{(\frac{M}{2})}$, upsample the disparity again, and apply the update $\frac{M}{2}$ more times to obtain $\hat D^{(M)}$ at resolution $(\frac{H}{4},\frac{W}{4})$.
Finally, we upsample the predicted disparity $\hat D^{(M)}$ to $(H,W)$.
Different versions of the function $g$ are trained for each of the three resolutions and upsampling uses the same mechanism as in RAFT.

The detailed structure of the update network $g$ is illustrated in \cref{fig:update_block}a and described below.

\paragraph{Correlation volume.}

Similar to classical disparity estimation methods, $g$ starts by computing the correlation volumes between left and right features $C_{t,s,k} = \mathrm{corr}(\phi_{t,s,k}^L, \phi_{t,s,k}^R) \in \mathbb{R}^{\frac{H}{sk} \times \frac{W}{sk} \times \frac{W}{sk}}$ for each feature resolution $k$ and at different scales $s \in \{1,2,4,8 \}$.
The correlation is computed along epipolar lines, which correspond to image rows as the input images are rectified.
Each element $(h, w, w')$ of $C_{t,s,k}$ is the inner product between feature vectors of $\phi_{t,s,k}^L$ and $\phi_{t,s,k}^R$:
\begin{equation}
    C_{t,s,k}(h, w, w')
    = \frac{1}{\sqrt{d}}\langle 
    \phi_{t,s,k}^L(h, w), \phi_{t,s,k}^R(h,w') 
    \rangle .
\end{equation}
Thus, $C_{t,s,k}(h, w, w')$ is proportional to how well the left image point $(h, w)$ matches the right image point $(h,w')$.
As the correlation volume does not depend on the update iteration $m$, it is computed only once.

\paragraph{Correlation lookup.}%
\label{lookup}

The function $g$ updates the disparity at location $(h,w)$ by observing the correlation volume in a local neighbourhood centered on the current disparity value $\hat{D}^{(m)}_t(h,w)$.
The necessary samples are collected and stacked as feature channels of the tensor
\begin{equation}
\hat{C}^{(m)}_{t,k}(h,w)
\!=\! \underset{s, \delta}{\mathrm{cat}}\!
\left[ 
C_{t,s,k}\!
\left(
    \frac{h}{s}, \frac{w}{s}, \frac{w\!+\!\hat{D}^{(m)}_t(h,w)}{s}\!+\!\delta
\right) 
\!\right]\!.
\end{equation}
The current correlation estimate $\hat{C}^{(m)}_{t,k} \in \mathbb{R}^{ \frac{H}{k} \times \frac{W}{k}\times 4(2\Delta+1)}   $ captures the correlation volume across all four scales and in a neighborhood $\delta \in \{-\Delta, \ldots, \Delta \}$ for additional context, with $\mathrm{cat}$ representing the concatenation operation.

\paragraph{Modality fusion.}
The estimated correlation, disparity, and feature maps need to be combined to provide the update block with enough information to update the disparity.
Using a late fusion scheme, we encode correlation $\hat{C}^{(m)}_{t,k}$ and disparity $\hat{D}^{(m)}_t$ separately, before concatenating them with the feature map of the left frame $\phi^L_{t,k}$, as it is the reference frame.
 To incorporate temporal and spatial information, we apply self-attention across time $(T)$ and space $(\frac{H}{k},\frac{W}{k})$ to the output of the modality fusion step.  For efficiency, we do it only for $k=16$. 
Architecture details can be found in \cref{fig:update_block} and the supplementary material.

\paragraph{3D CNN-based GRU.}
The update function $g$ is implemented using a 3D convolutional GRU.

At each step $m$, the GRU takes as input the fused features and a hidden state that is initialized with the reference feature map $\phi^{L}$.
All internal operations of the GRU are implemented as separable 3D convolutions across space and time to propagate temporal and spatial information.
The output of each iteration is an update to the current disparity estimate as in \cref{eq:update}.
Each subsequent iteration is preceded by correlation lookup and modality fusion.

\begin{table*}[t]

\setlength\tabcolsep{.4em}
\centering
\begin{tabular}{lllll}
 \includegraphics[width=.17\linewidth,valign=m]{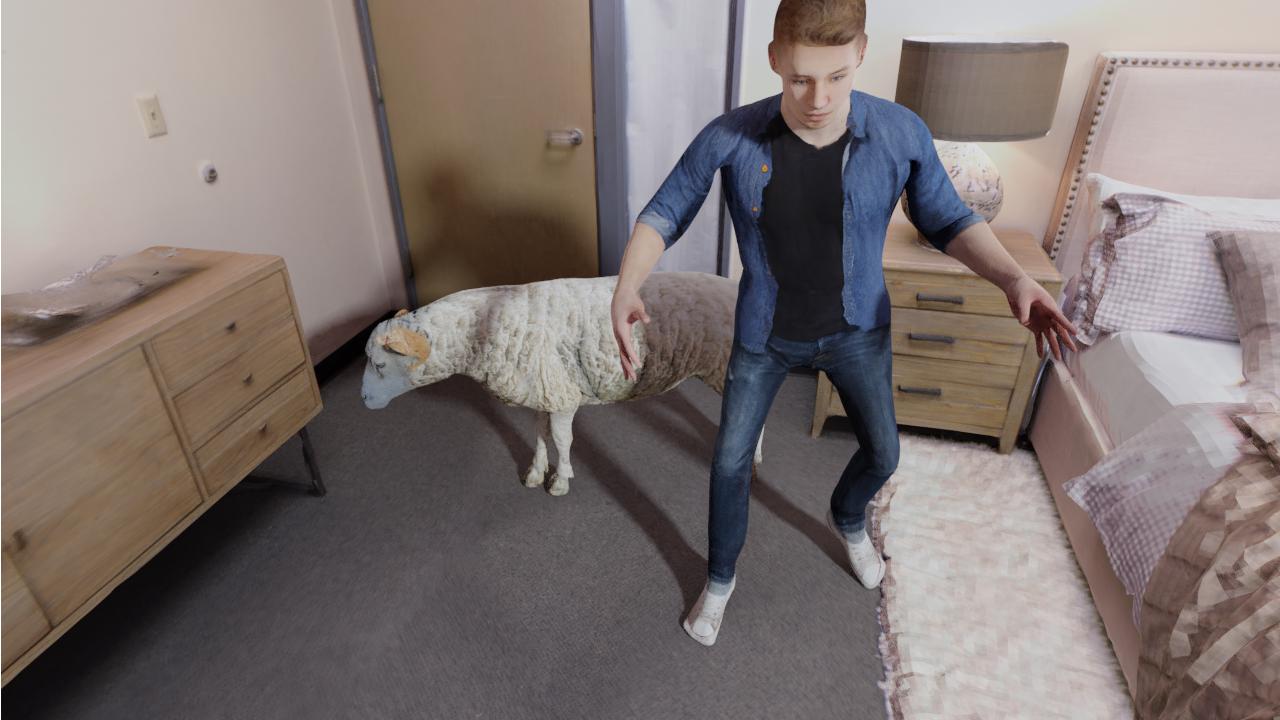} & \includegraphics[width=.17\linewidth,valign=m]{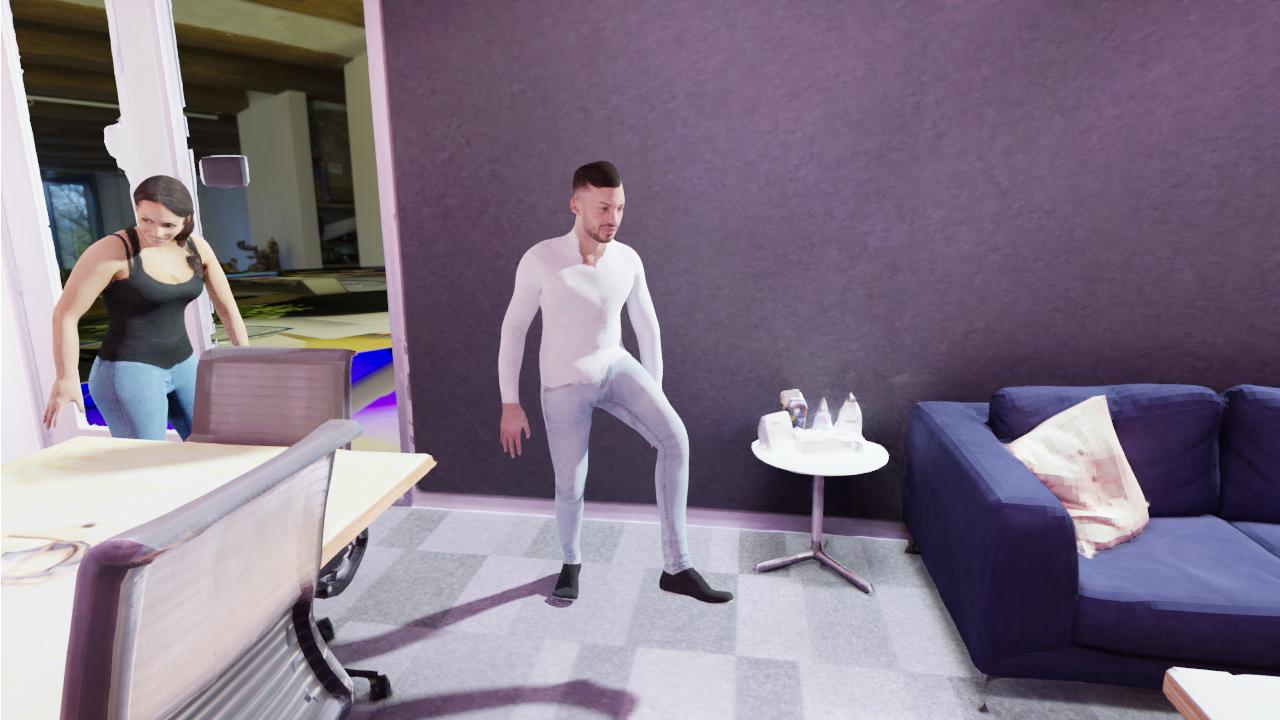} & \includegraphics[width=.17\linewidth,valign=m]{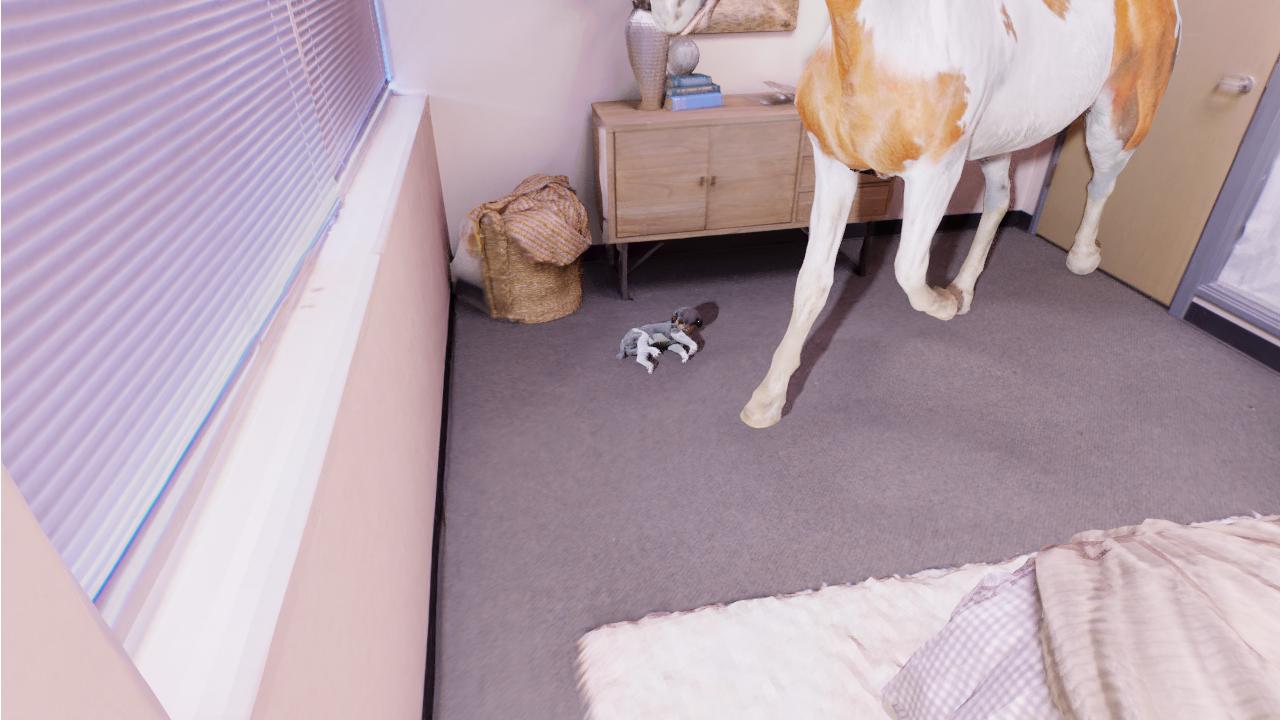}& \includegraphics[width=.17\linewidth,valign=m]{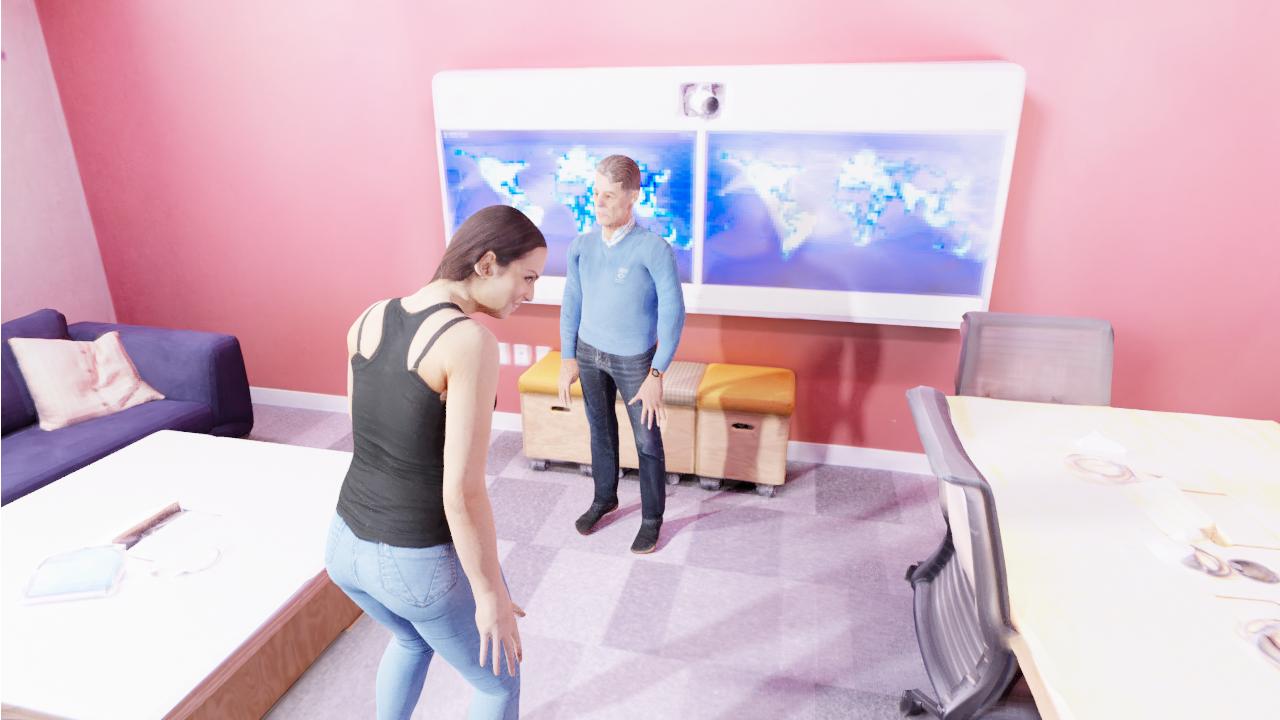}& \includegraphics[width=.17\linewidth,valign=m]{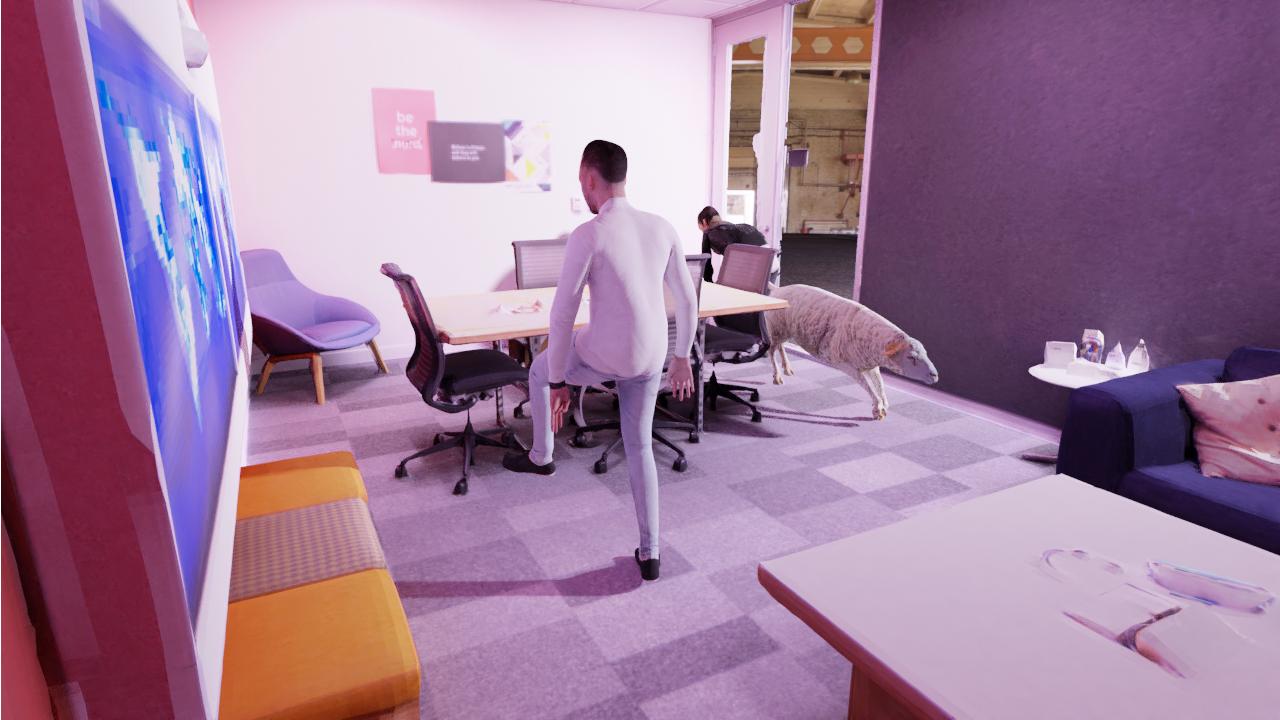}\\
 \includegraphics[width=.17\linewidth,valign=m]{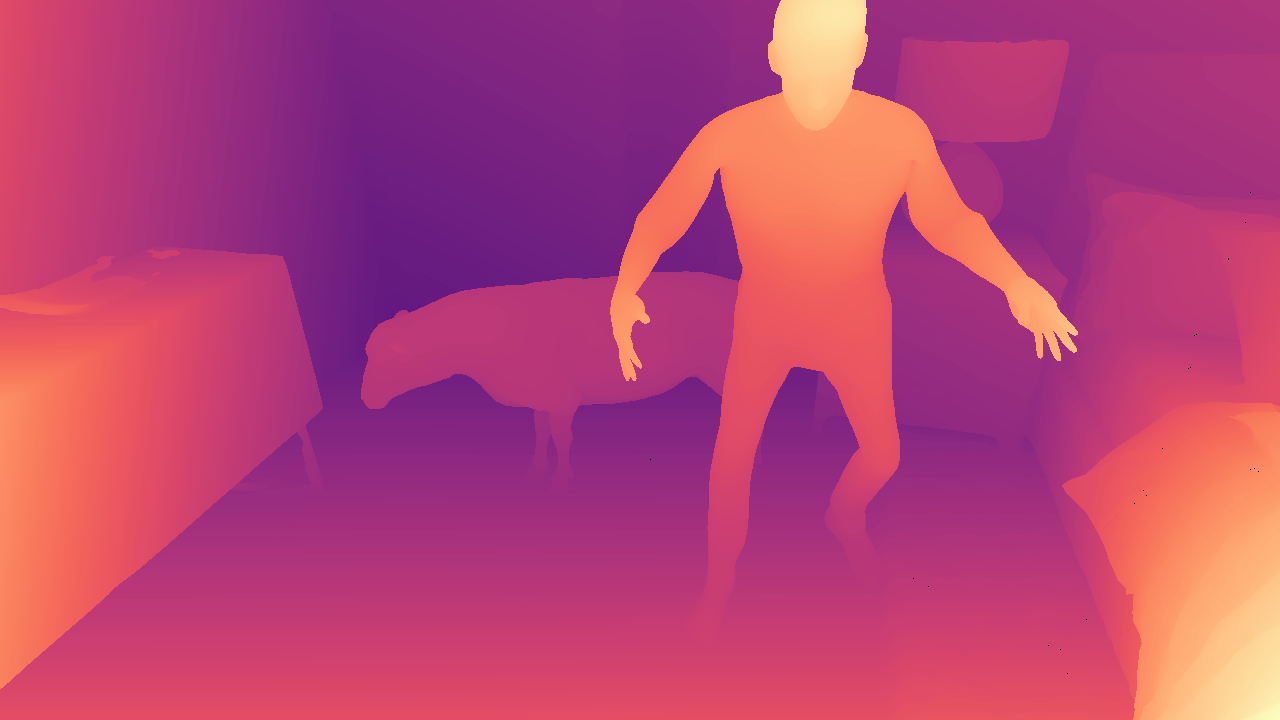} & \includegraphics[width=.17\linewidth,valign=m]{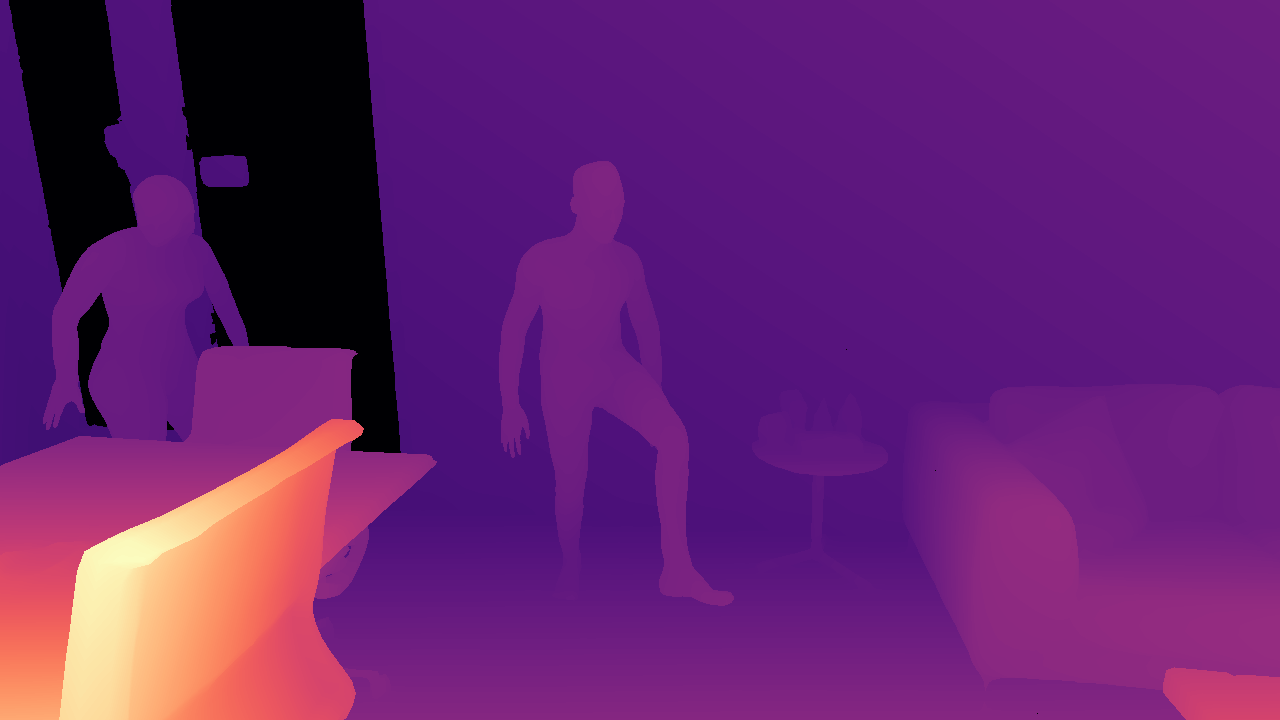} & \includegraphics[width=.17\linewidth,valign=m]{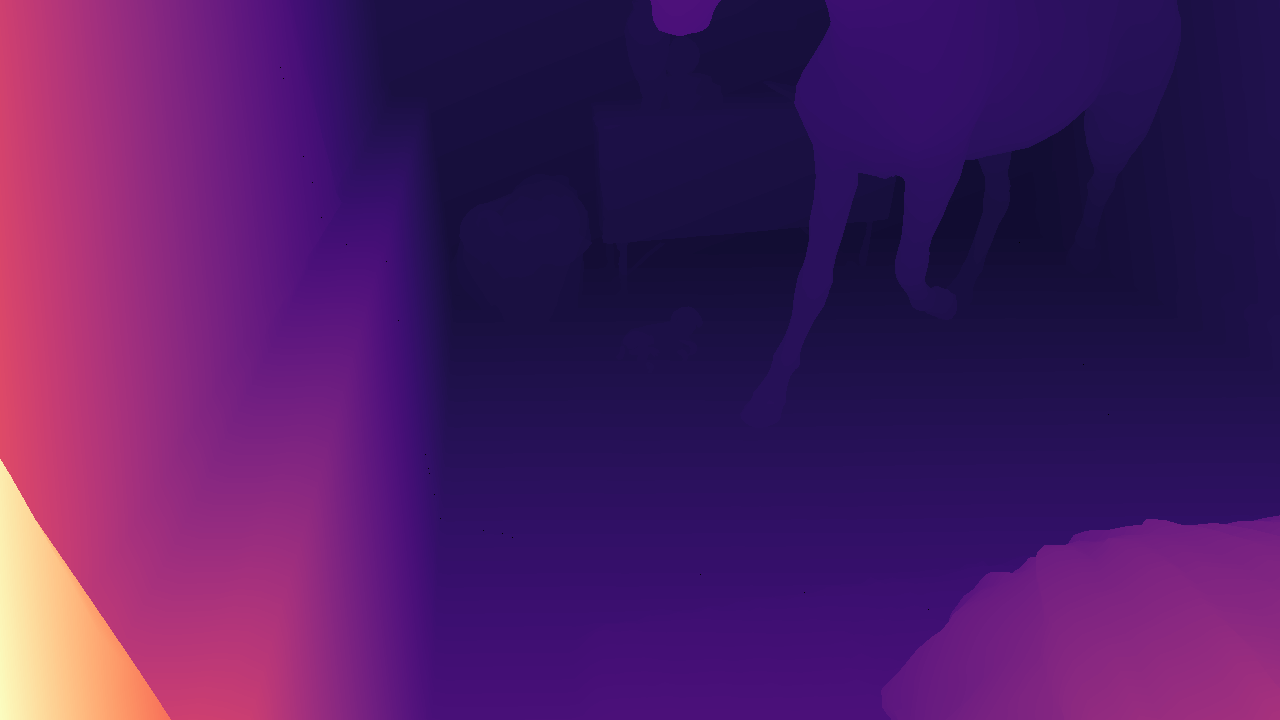}& \includegraphics[width=.17\linewidth,valign=m]{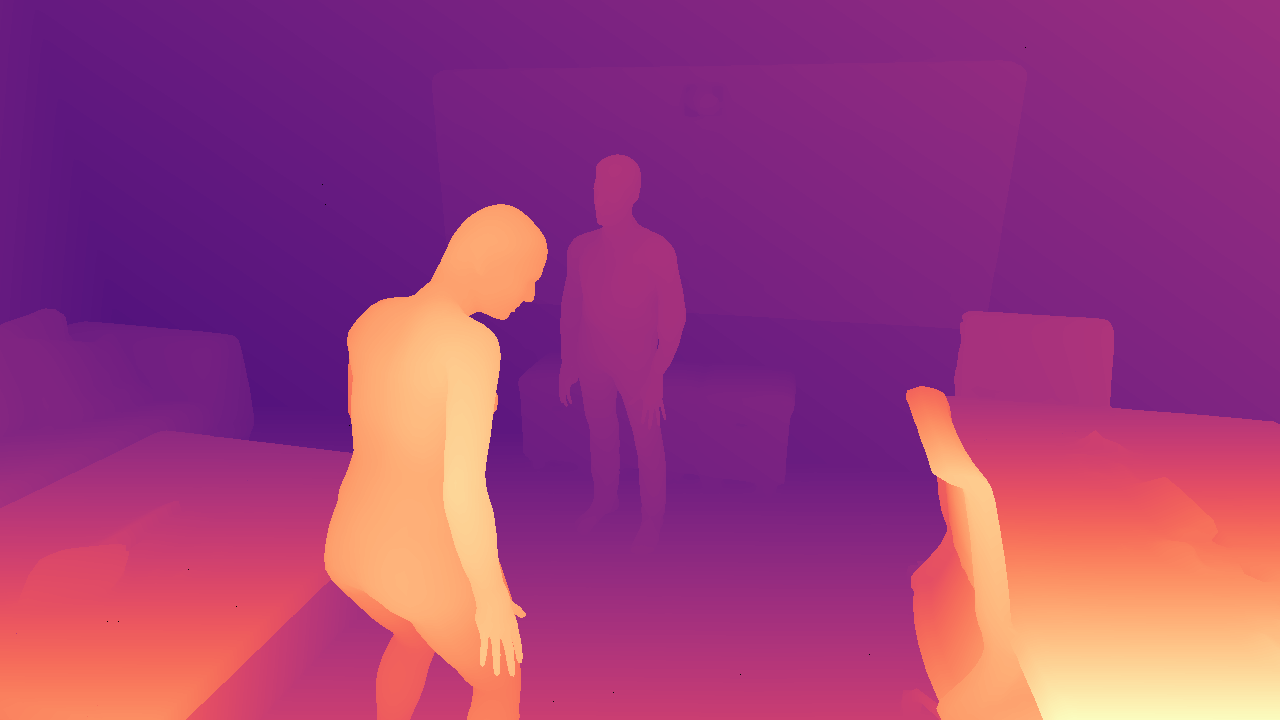}& \includegraphics[width=.17\linewidth,valign=m]{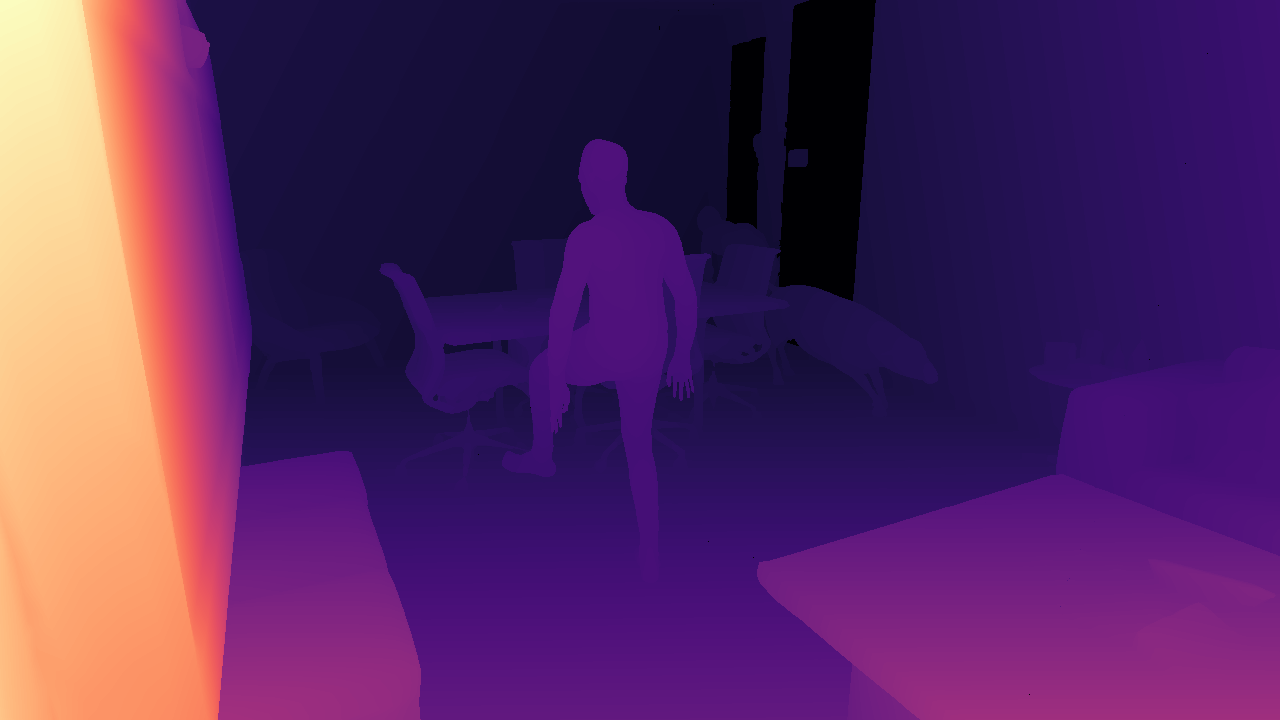}\\
\includegraphics[width=.17\linewidth,valign=m]{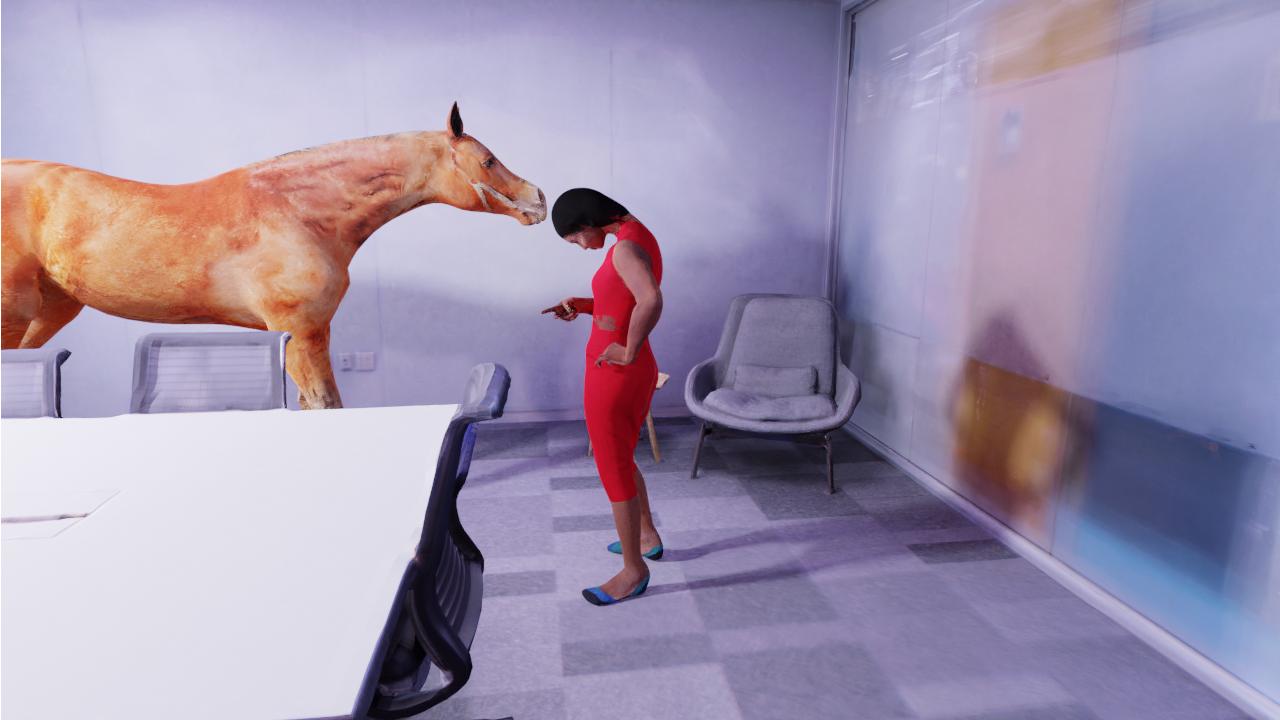} & \includegraphics[width=.17\linewidth,valign=m]{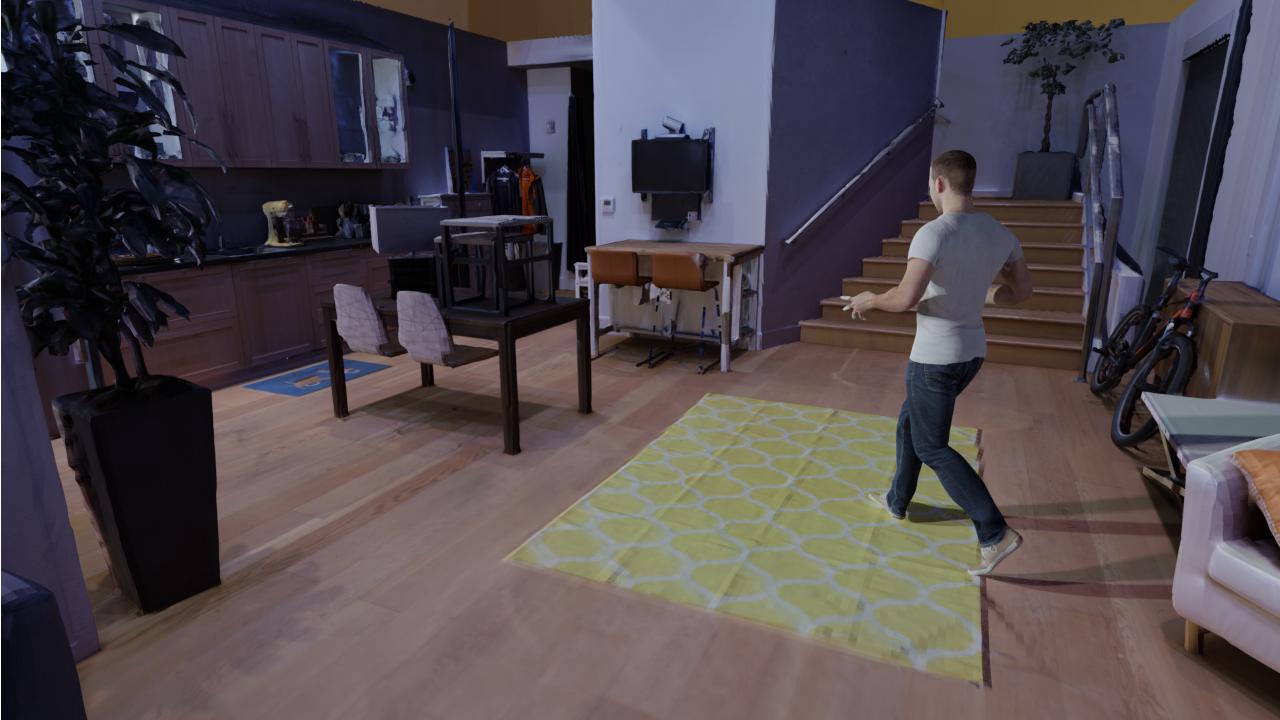} & \includegraphics[width=.17\linewidth,valign=m]{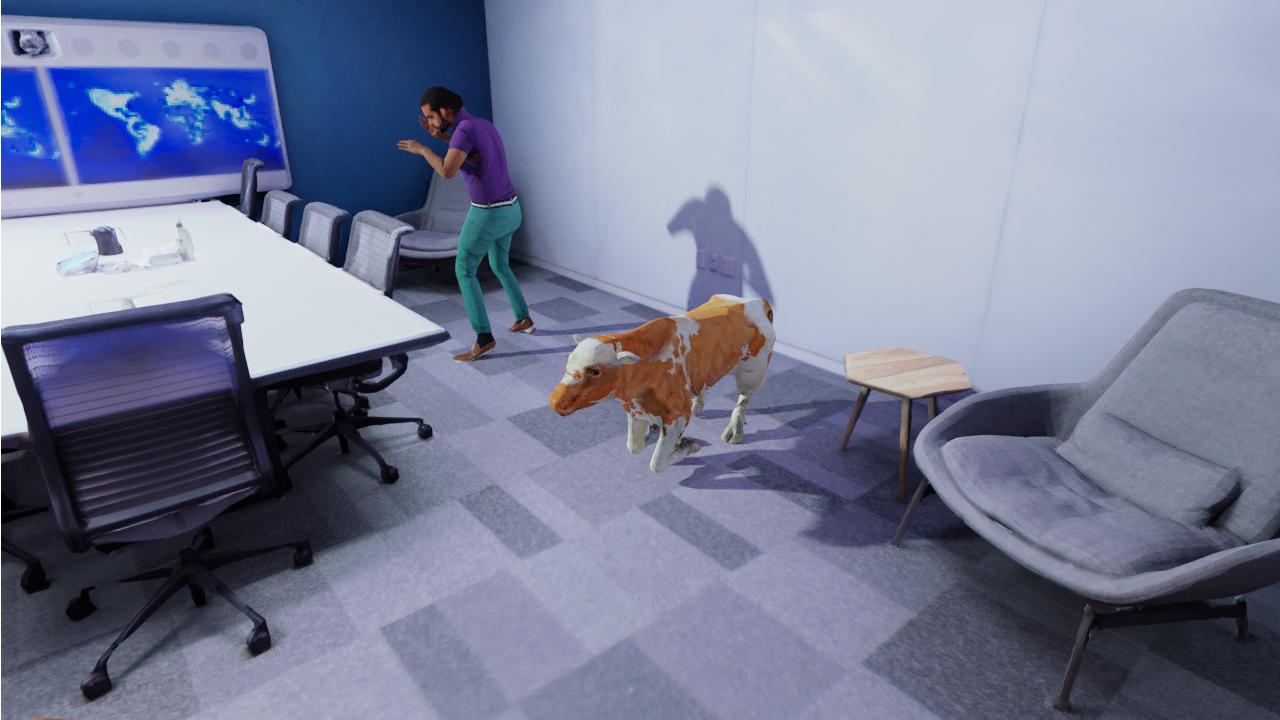}& \includegraphics[width=.17\linewidth,valign=m]{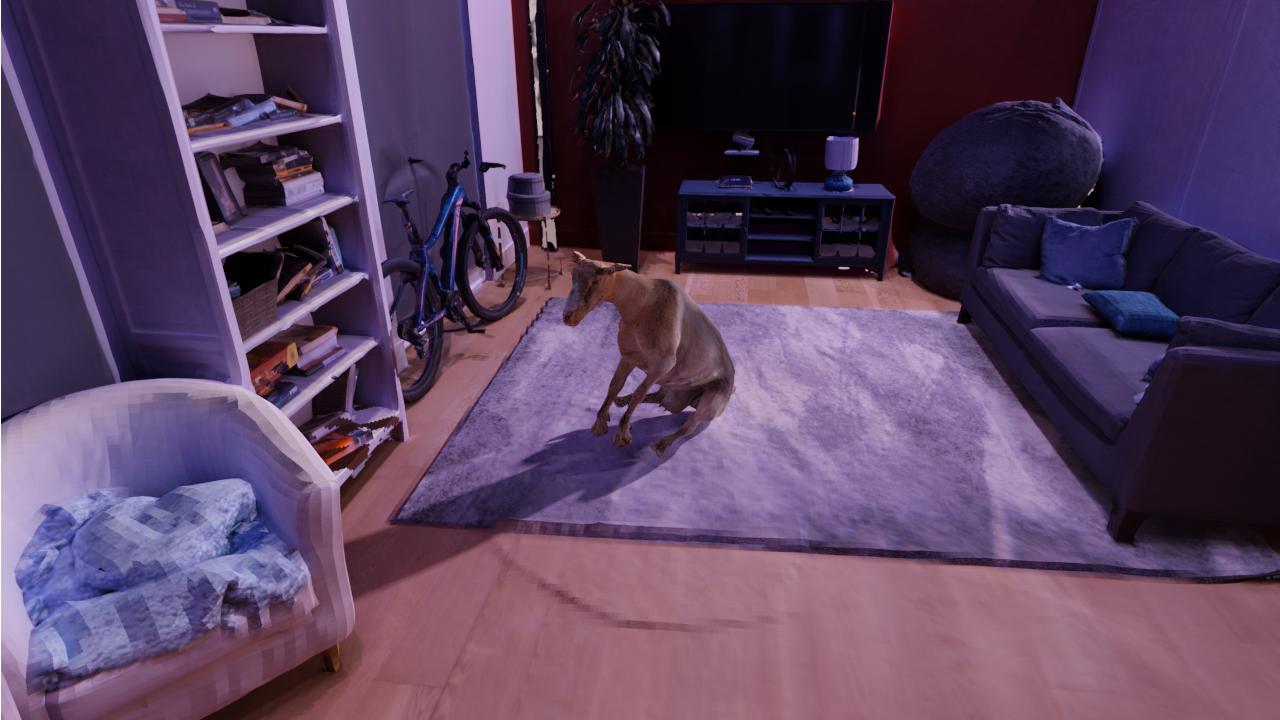}& \includegraphics[width=.17\linewidth,valign=m]{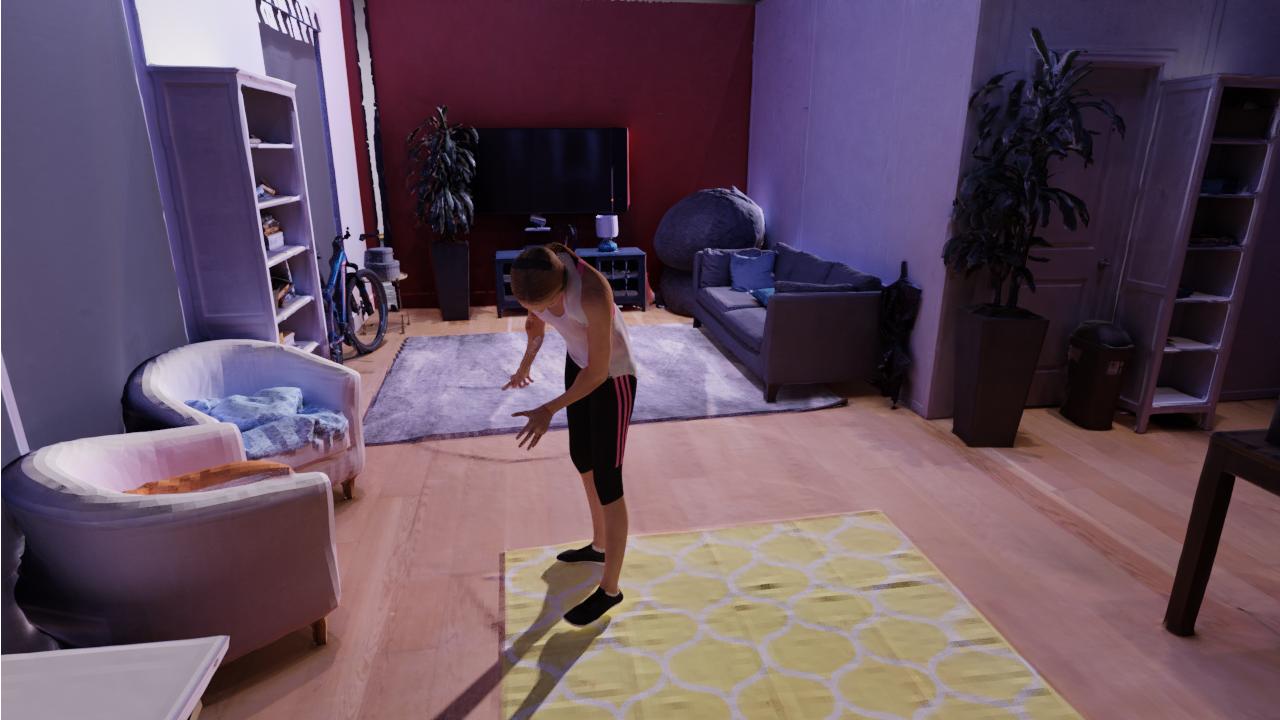}\\
\includegraphics[width=.17\linewidth,valign=m]{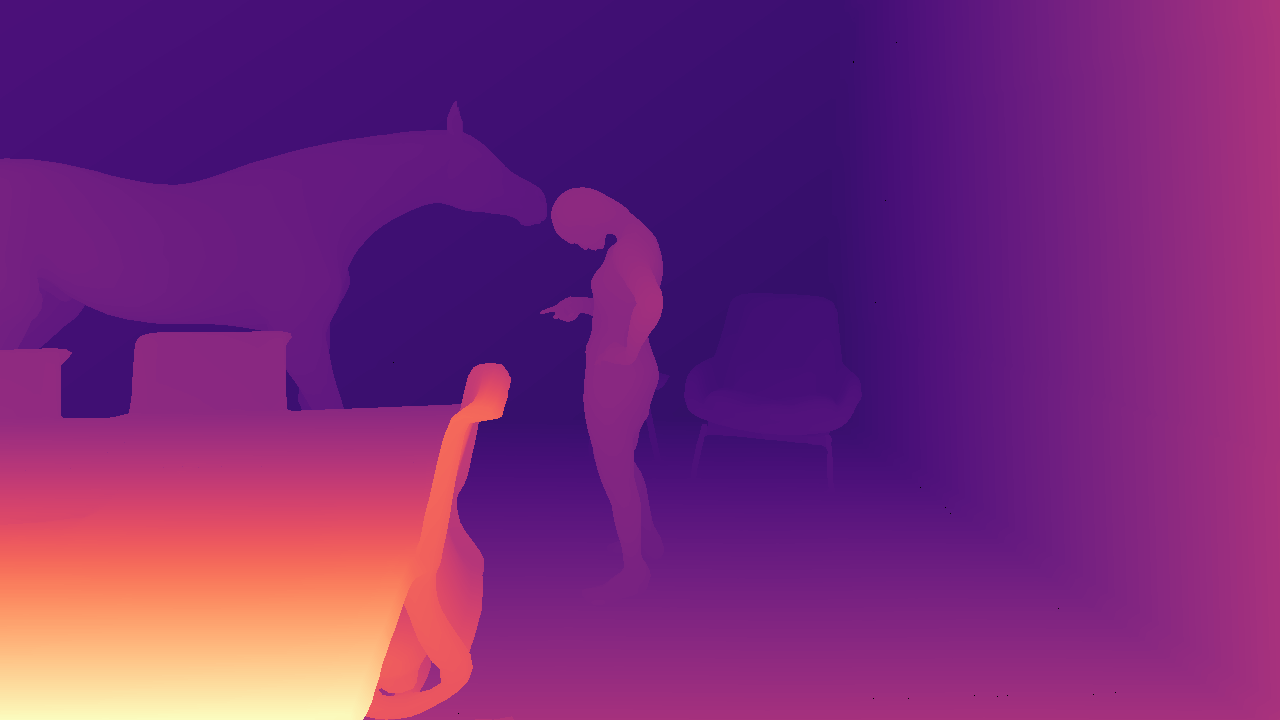} & \includegraphics[width=.17\linewidth,valign=m]{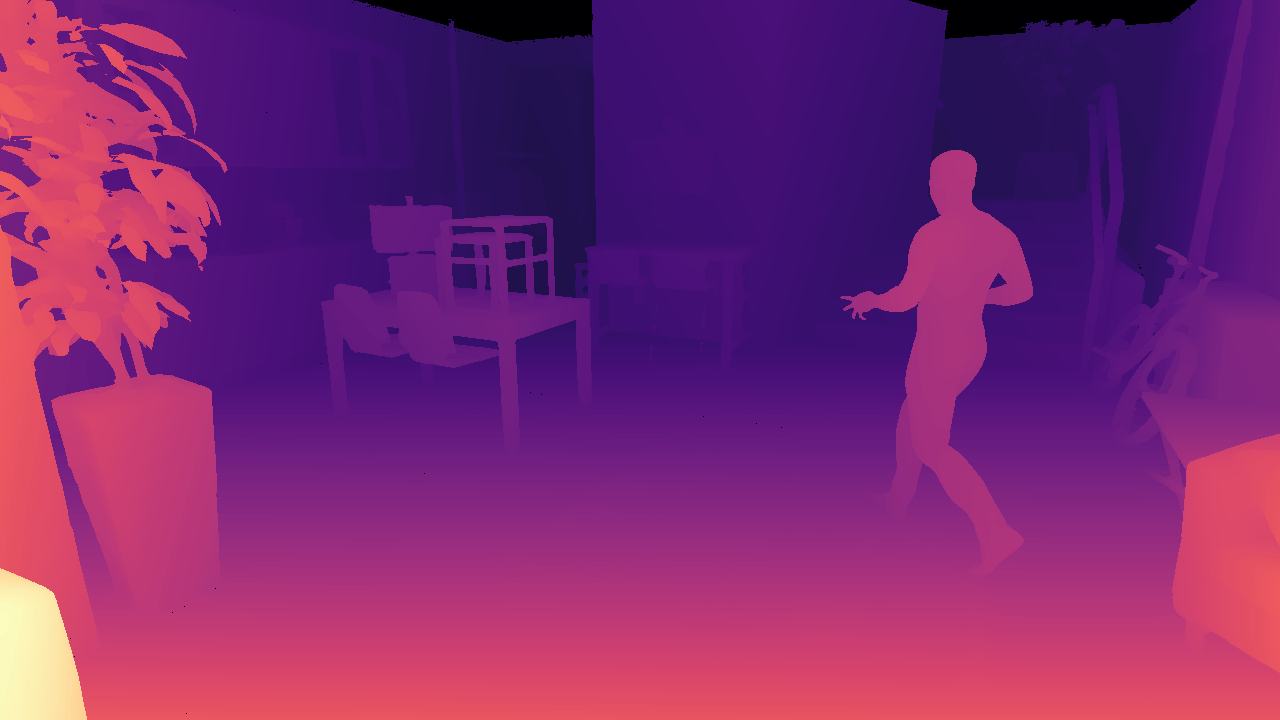} & \includegraphics[width=.17\linewidth,valign=m]{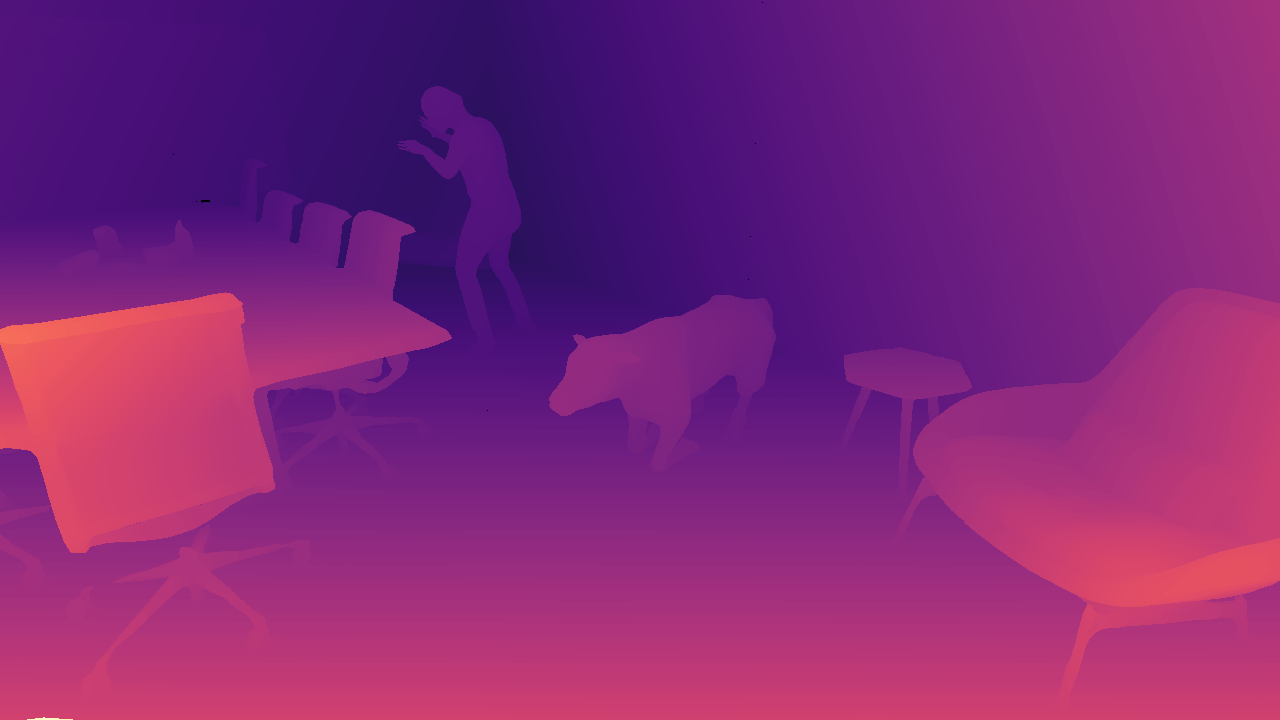}& \includegraphics[width=.17\linewidth,valign=m]{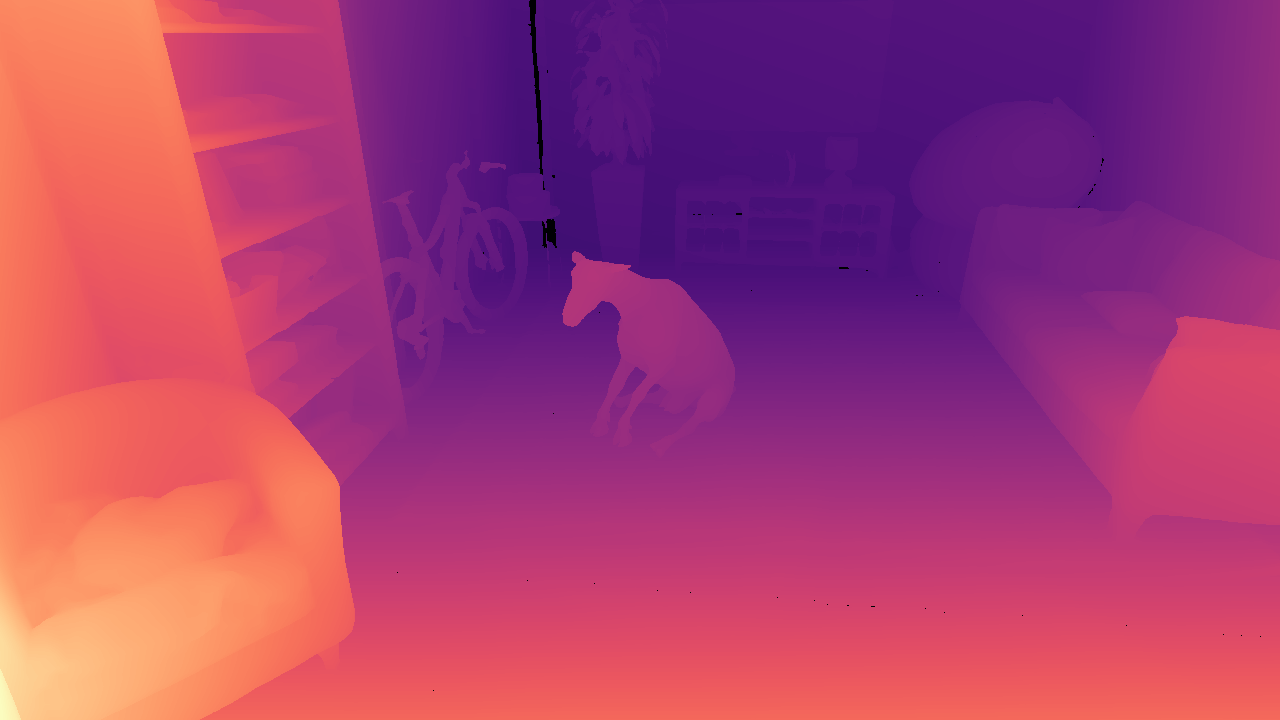}& \includegraphics[width=.17\linewidth,valign=m]{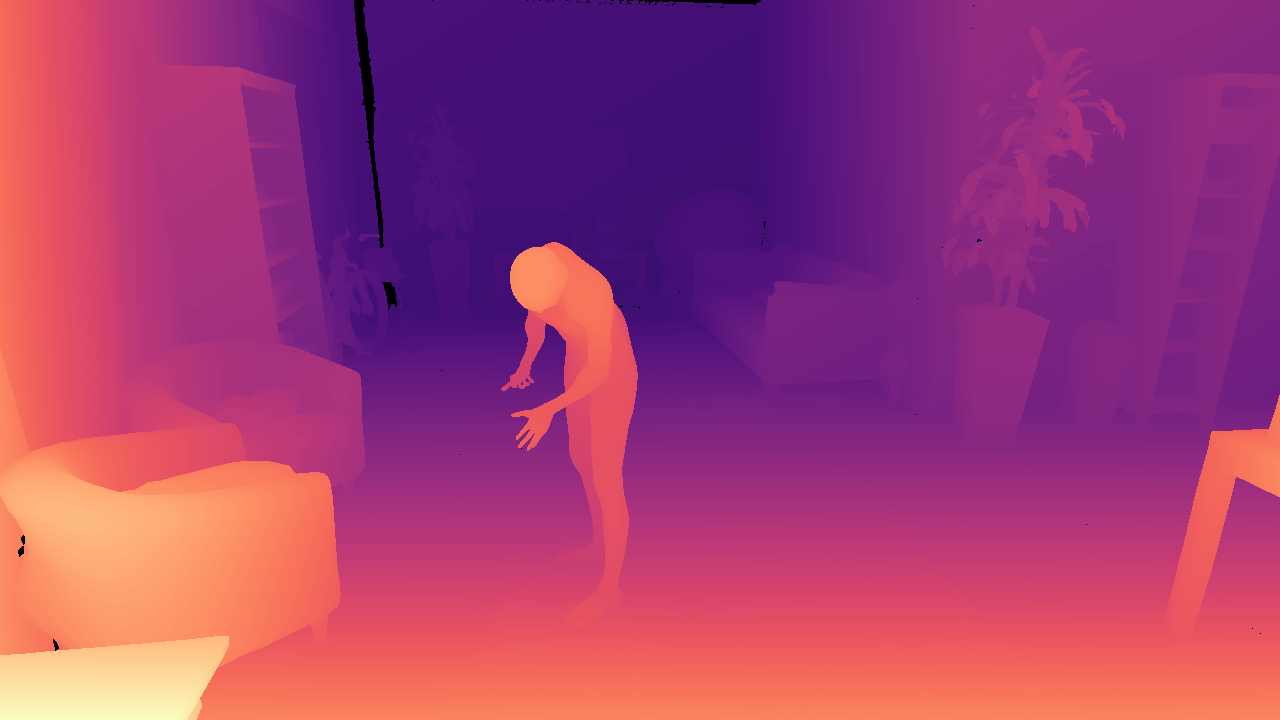}\\
\end{tabular}
\vspace{1mm}
\caption{\textbf{Dynamic Replica.} Example video frames and disparity maps from our proposed dataset. The proposed dataset contains diverse indoor scenes filled with animated people and animals to provide a benchmark for disparity estimation in close-to-real environments.}%
\label{tab:images_dynamic_replica}
\end{table*}

\subsection{Training}
Due to its iterative nature, the proposed model generates $M$ predictions for every timestamp.
During training, we supervise the network over the full sequence of predictions $\hat{D}_t^{(m)}$, with exponentially increasing weights towards the final estimate at step $M$ as follows:

\begin{equation}
    \mathcal{L}(\hat{D}, D) = \sum_{t=1}^{T} { \sum_{m=1}^{M} \gamma^{M-m} \|\hat{D}_t^{(m)}  - D_t\|},
  \label{eq:also-important}
\end{equation}
where $\gamma=0.9$ and $D$ is the ground truth disparity sequence. Lower resolution disparity estimates are up-sampled to ground truth resolution. 

\subsection{Inference}
The model is trained on stereo sequences with a fixed length of $T$. To increase the temporal coherence at test time, we apply it to videos of arbitrary length using a sliding window approach with overlap, where we discard the predictions in overlapping time steps. For details see supp. mat.
Videos shorter than $T$ time steps can be used by repeating the first or the last frame up to the minimal length $T$. This corresponds to a static scene without moving camera and is well within the training distribution of the model.

\section{Dynamic Replica: dynamic stereo dataset}%
\label{sec:dr}

Training temporally consistent models requires a dataset with stereo videos and dense ground-truth annotations. 
While the SceneFlow dataset \cite{MIFDB16} fulfils both criteria, it is comprised of very short sequences of randomly textured objects moving.
In this paper, we introduce a more realistic synthetic dataset and benchmark of animated humans and animals in every-day scenes: \emph{Dynamic Replica}.

The dataset consists of $524$ videos of synthetic humans and animals performing actions in virtual environments (see \cref{tab:images_dynamic_replica}). Training and validation videos are $10$ seconds long and each contain $300$ frames. There are $484$ training and $20$ validation videos.
Our test split consists of $20$ videos of length $30$ seconds to benchmark models on longer videos.

Videos are rendered at a resolution of $1280\!\times\!720$. This is close to the resolution of modern screens and higher than the resolution of such popular stereo datasets as Sintel ($1024\!\times\!536$) and SceneFlow ($960\!\times\!540$).

The dataset is based on Facebook Replica~\cite{replica19arxiv} reconstructions of indoor spaces. We take $375$ 3D scans of humans from the RenderPeople\footnote{\href{http://renderpeople.com/}{http://renderpeople.com/}} dataset and animate them using motion capture sequences from real humans scans. We use artist-created animated 3D models of animals from $13$ categories (chimp, dog, horse, sheep, etc.). We use different environments, scans, textures and motions for training, validation and test splits to evaluate generalization to unseen environments and dynamic objects.

 We randomize camera baselines in our training subset to ensure generalization across different stereo setups. Baselines are sampled uniformly between 4cm and 30cm.
  
 For each scene we generate a camera trajectory imitating a person filming the scene with their mobile phone or AR glasses. These virtual cameras have smooth trajectories and are located approximately at $1.25$m above the ground.

All samples in the dataset contain ground-truth depth maps, optical flow, foreground / background segmentation masks and camera parameters for both stereo views.

\section{Experiments}

We structure the experiments as follows.
First, we evaluate the \emph{Dynamic Replica} dataset by comparing generalization performance of prior models trained on other datasets and on \emph{Dynamic Replica}. 
Then we evaluate our model and compare it to the state of the art in temporal consistency.
Finally, we ablate design choices in the model architecture and verify the importance of its individual components.

\paragraph{Implementation Details.}
\label{impl}
We implement \emph{DynamicStereo} in PyTorch~\cite{paszke2019pytorch} and train on $8$ NVIDIA TESLA Volta V100 32GB GPUs.
The SF version is trained for $70$k iterations with a batch size $8$. We train the DR+SF version for $120$k iterations which takes about $4$ days. 

Both models are trained on random $384\!\times\!512$ crops of sequences of length $T=5$ and evaluated in the original resolution with $T = 20$ and overlapping windows of size $10$.
During training, we use the AdamW optimizer~\cite{loshchilov2017decoupled} and set the number of iterative updates $M=20$.
We train with one-cycle learning rate schedule \cite{smith2019super} with a maximum learning rate $3 \cdot 10^{-4}$. We set the lookup neighborhood $\Delta=4$ (see \cref{lookup}).
For attention layers we use positional encoding in both space and time. We apply linear attention \cite{sun2021loftr} for space 
and use standard quadratic attention for time.
For other implementation details, please see supp. mat.

\subsection{Dynamic Replica}
\setlength\tabcolsep{.15em}
\begin{table}[t]
\centering
\footnotesize
\begin{tabular}{ llccccccccc }
\toprule
\multirow{2}{*}{Mtd} & \multirow{2}{*}{Data} & \multirow{2}{*}{KITTI} & \multicolumn{3}{c}{Middlebury} & \multirow{2}{*}{ETH3D} & \multicolumn{2}{c}{Sintel Stereo} & DR \\
\cmidrule(lr){4-6}
& & & full & half & quarter & & Clean & Final & full \\
\midrule
\midrule
\cite{raft_stereo} & DR & 7.25 & 19.51 & 13.13 & 13.86 & 3.78 & 7.36 & 13.04 & 2.90 \\
\cite{raft_stereo} & SF & \textbf{5.55} & 17.76 & 12.80 & 9.64 & 3.05 & \textbf{5.89} & 9.20 & 4.01\\
\cite{raft_stereo} & (DR+SF)/2 & 5.63 & \textbf{15.08} & \textbf{10.36} & \textbf{9.24} & \textbf{3.02} & 5.96 & \textbf{9.12} & \textbf{2.20}\\
\midrule
\cite{CREStereoLi2022practical}
& DR & 6.04 & 31.94 & 23.82 & 15.93 & 3.94 & 15.03 & 19.35 & 4.13 \\
\cite{CREStereoLi2022practical} & SF & 6.22 & 23.95 & 15.64 & 10.50 & 3.95 & 7.51 & 11.01 & 6.59 \\
 \cite{CREStereoLi2022practical} & (DR+SF)/2 & \textbf{5.35} & \textbf{22.02} & \textbf{13.69} & \textbf{8.96} & \textbf{3.53} & \textbf{6.96} & \textbf{9.98} & \textbf{3.49} \\
\midrule
\end{tabular}
\vspace{1mm}
\caption{\textbf{Dynamic Replica generalization -- non-temporal disparity estimation. } SF - SceneFlow \cite{Mayer_2016}, DR - \textit{Dynamic Replica} (ours). For (DR+SF)/2, we replace half of SF with samples from DR. The performance improves over pure SF training, showing that DR is valuable for generalization, compared to training only with SF data. Errors are the percent of pixels with end-point-error greater than the specified threshold.  We average across 3 runs with different seeds and use the standard evaluation thresholds: 3px for KITTI 2015 \cite{Kitti} and Sintel Stereo \cite{Sintel}, 2px for Middlebury \cite{scharstein2014high}, 1px for ETH3D \cite{schops2017multi} and DR.}%
\label{tab:dataset_comparison}
\end{table}
In \cref{tab:dataset_comparison} we show results obtained by training two recent state-of-the-art models, RAFT-Stereo \cite{raft_stereo} and CRE-Stereo \cite{CREStereoLi2022practical} on SceneFlow, our dataset and their combination and testing the models' generalization to other disparity estimation datasets. While the main objective of our method is to produce temporally consistent disparity estimates, here we train state-of-the-art disparity estimation models to evaluate the usefulness of the dataset in this setting.

SceneFlow is an abstract dataset of moving shapes on colorful backgrounds. 
It aims at generalization through domain randomization, while our dataset consists of more realistic home and office scenes with people.
Thus, while training on \emph{Dynamic Replica} alone improves performance on the DR test set, it does not generalize as well as models trained on SceneFlow. However, combining both datasets boosts performance across both, datasets and models.

This highlights the benefits of including \emph{Dynamic Replica} in the standard set of disparity estimation training datasets, even if its main goal is to enable training of temporally consistent models on longer sequences. 

\begin{table}[t]
\setlength\tabcolsep{.3em}
\centering
\begin{tabular}{ccc}
 \includegraphics[width=.32\linewidth,valign=m]{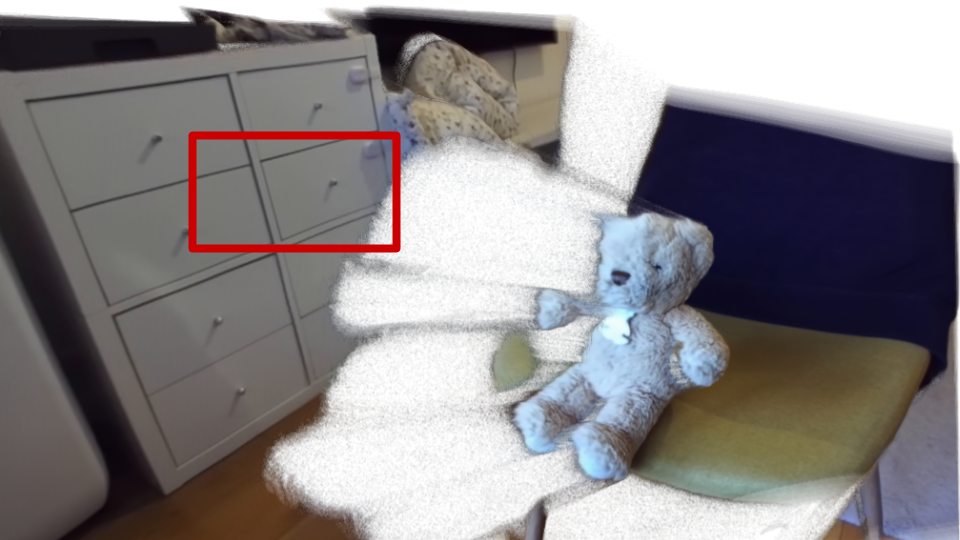}  & \includegraphics[width=.32\linewidth,valign=m]{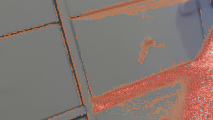} & \includegraphics[width=.32\linewidth,valign=m]{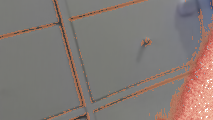}  \\
  \includegraphics[width=.32\linewidth,valign=m]{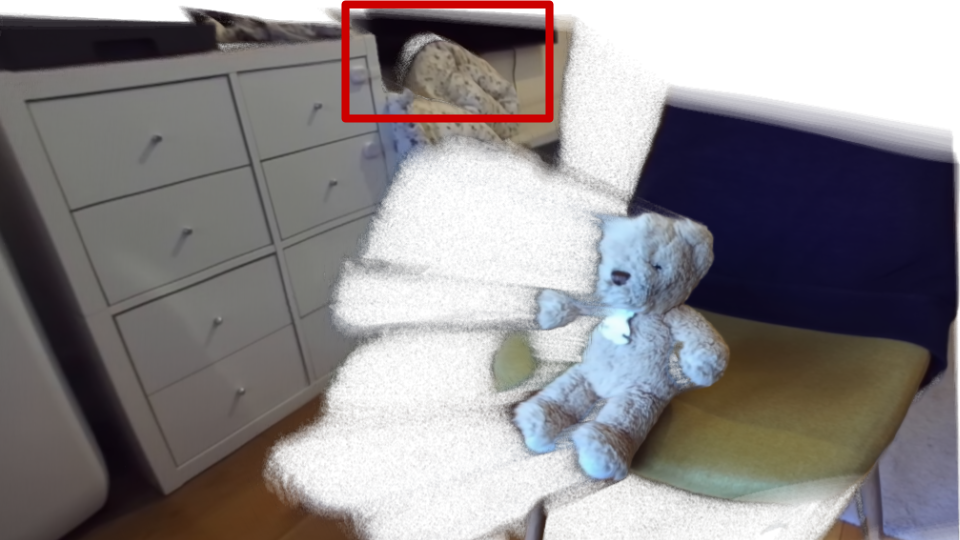}  &
  \includegraphics[width=.32\linewidth,valign=m]{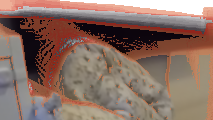}
  &
  \includegraphics[width=.32\linewidth,valign=m]{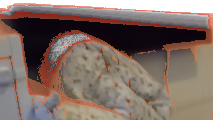} \vspace*{4pt}\\
 \small (mean image) & \small RAFT Stereo~\cite{raft_stereo} & \textbf{\small DynamicStereo} \\
\end{tabular}
\caption{\textbf{Temporal consistency.} Mean and variance of a 40-frame reconstructed static video, visualized. Both models are trained on DR \& SF. We predict depth for each frame and convert it to globally aligned point clouds. Render combined point cloud with a camera displaced by 15 angles. Finally, we compute mean and variance across all images. Pixels with variance higher than 50px$^2$ are shown in red. Our method has lower variance. }
\label{fig:temporal_consistency}
\end{table}

\subsection{Temporal Consistency}
\label{tepe}
\begin{table*}[t]
\centering
\begin{tabular}
{clc|ccc @{\hspace{10\tabcolsep}} c|ccc @{\hspace{10\tabcolsep}}c|ccc}
\toprule
\multirow{4}{*}{Training data} & \multirow{4}{*}{Method} & \multicolumn{8}{c}{Sintel Stereo} & \multicolumn{4}{c}{Dynamic Replica} \\
\cmidrule(lr){3-10}\cmidrule(lr){11-14}
& & \multicolumn{4}{c}{Clean} & \multicolumn{4}{c}{Final} &  \multicolumn{4}{c}{First 150 frames}  \\
\cmidrule(lr){3-6}\cmidrule(lr){7-10}\cmidrule(lr){11-14}
& & \small{\manconcentriccircles} & \multicolumn{3}{c}{\large{\clock}} & \small{\manconcentriccircles} & \multicolumn{3}{c}{\large{\clock}} & \small{\manconcentriccircles} & \multicolumn{3}{c}{\large{\clock}} \\
& & \phantom{-}$\delta_{3px}$\phantom{-} & TEPE & $\delta^t_{1px}$ & $\delta^t_{3px}$ & \phantom{-}$\delta_{3px}$\phantom{-} & TEPE  & $\delta^t_{1px}$ & $\delta^t_{3px}$ &  \phantom{-}$\delta_{1px}$\phantom{-} & TEPE  & $\delta^t_{1px}$ & $\delta^t_{3px}$ \\

\midrule
\multirow{3}{*}{SF} &  CODD \cite{Holograms}  & 8.68 & 1.44 & 10.78 & 5.65 & 17.46 & 2.32 & 18.56 & 9.79 & 6.59 & 0.105 & 1.04 & 0.42 \\
& RAFT-Stereo \cite{raft_stereo} & 6.12 & 0.92 & 9.33 & 4.51 & 10.40 & 2.10 & 13.69 & 7.08 & 5.51 & 0.145 & 2.03 & 0.65 \\
& Ours & \textbf{6.10} & \textbf{0.77} & \textbf{8.41} & \textbf{3.93} & \textbf{8.97} & \textbf{1.45} & \textbf{11.95} & \textbf{5.98} & \textbf{3.44}  & \textbf{0.087} & \textbf{0.75} & \textbf{0.24} \\
\midrule
\multirow{2}{*}{DR+SF} &  RAFT-Stereo \cite{raft_stereo} & \textbf{5.71} & 0.84 & 9.15 & 4.40 & 9.16 &  2.27 & 13.45 & 7.17 & \textbf{1.89} & \textbf{0.075} &  0.77 & 0.25  \\
& Ours & 5.77 & \textbf{0.76} & \textbf{8.46} & \textbf{3.93} & \textbf{8.68} & \textbf{1.42} & \textbf{11.93} & \textbf{5.92} & 3.32 & \textbf{0.075} & \textbf{0.68} &  \textbf{0.23} \\
\midrule
SF+M+K &  CODD \cite{Holograms} & 9.11 & 1.33 & 12.16 & 6.23 & 11.90 & 2.01 & 16.16 & 8.64 & 10.03 & 0.152 & 2.16 & 0.77 \\
SF+M & RAFT-Stereo \cite{raft_stereo} & 5.86 & 0.85 & 8.79 & 4.13 & \textbf{8.47} & 1.63 & 12.40 & 6.23 & 3.46 & 0.114 & 1.34 & 0.41  \\
 \phantom{-}7 datasets (incl. Sintel) \phantom{-} & CRE-Stereo \cite{CREStereoLi2022practical} & \textcolor{gray}{4.58} &  \textcolor{gray}{0.67} & \textcolor{gray}{6.36} & \textcolor{gray}{3.26} & \textcolor{gray}{8.17} & \textcolor{gray}{1.90} & \textcolor{gray}{12.29} & \textcolor{gray}{6.87} & \textbf{1.75} & 0.088 & 0.88 & 0.29 \\
DR+SF & Ours & \textbf{5.77} & \textbf{0.76} & \textbf{8.46} & \textbf{3.93} & 8.68 & \textbf{1.42} & \textbf{11.93} & \textbf{5.92} & 3.32 & \textbf{0.075} & \textbf{0.68} & \textbf{0.23}  \\
\bottomrule
\end{tabular}
\caption{\textbf{Accuracy \small{\manconcentriccircles} and Temporal consistency \large{\clock}.} SF - SceneFlow \cite{Mayer_2016}, K - KITTI \cite{Kitti}, M - Middlebury \cite{scharstein2014high}, DR - Dynamic Replica (ours). Temporal end-point error (TEPE) measures the consistency of the disparity estimation over time. We also compute $\delta^t_{1px}$ and $\delta^t_{3px}$ that show the proportion of pixels with TEPE higher than the threshold. Our model is more consistent than prior work and models. Additionally, other methods trained on our dataset also improve in temporal consistency. As CRE Stereo trains on Sintel and disparity estimation is evaluated on the training set of Sintel, CRE Stereo results are training set results and cannot be directly compared.} \label{tab:temporal_consitency}
\end{table*}

The main objective of our method and dataset is to enable training of temporally consistent disparity estimators. 
To evaluate temporal consistency, we compute the temporal end-point-error (TEPE) defined as follows:
\begin{equation}
    \mathrm{TEPE}(\hat{D}, D)\!=\!\!\sqrt {\sum_{t=1}^{T-1}\! \Big(\!(\hat{D_t}\!-\!\hat{D}_{t+1}) \!-\! (D_t\!-\!D_{t+1})\!\Big)^2\!\!.\!}
\end{equation}
This effectively measures the variation of the end-point-error across time. 
Lower values mean greater temporal consistency.
In \cref{tab:temporal_consitency} we show that our model is more temporally consistent than prior methods across training datasets and benchmarks. It is even better than CRE-Stereo~\cite{CREStereoLi2022practical} that is trained on a combination of seven datasets. 
Additionally, models trained with \emph{Dynamic Replica} are more consistent than models trained on SceneFlow alone.

With the above experiments, we have shown that \emph{Dynamic Replica} is a useful dataset for training temporally consistent disparity estimators as well as standard stereo-frame models, and that our proposed model improves over the state of the art in both tasks when trained on our dataset.

In \cref{fig:temporal_consistency}, we show a qualitative comparison of the temporal consistency between our model and RAFT-Stereo on a real-world sequence. We show the mean reconstruction over time of the sequence and color each pixel more red, the higher its variance across time. Our model significantly reduces the flickering that single-timestep models such as RAFT-Stereo produce.
For more qualitative results and videos, please see the supplementary material.

\subsection{Ablation Studies}
\label{sec:ablations}

In this section, we validate our method by ablating the design choices of our model.
We train the model on SceneFlow \cite{Mayer_2016} for $50$k iterations with the hyper-parameters described in Implementation Details (see Sec. \ref{impl}).
We evaluate these models on the clean pass of Sintel \cite{Sintel} and on the test split of \emph{Dynamic Replica}. We measure both accuracy and temporal consistency. For accuracy, we use an end-point-error threshold of $3$px for Sintel and $1$px for DR. This shows the proportion of pixels with an end-point-error higher than 3px. For consistency, we use TEPE (see Sec. \ref{tepe}).

\paragraph{Update Block.}
\begin{table}[t]
\centering
\begin{tabular}{ clcccc}
\toprule
\multirow{2}{*}{Method} & \multicolumn{2}{c}{Sintel Clean} & \multicolumn{2}{c}{Dynamic Replica} \\
\cmidrule(lr){2-3}\cmidrule(lr){4-5}
& \;$\delta_{3px}$\; & \;\;TEPE\;\; & \;$\delta_{1px}$\; & \;\;TEPE\;\; & \\
\midrule
\;\;shared weights\;\;  & 6.60 & 0.908  & 6.48 & \textbf{0.101}  \\
\textbf{sep. weights} & \textbf{6.24} & \textbf{0.823} & \textbf{4.76} & 0.126 \\
\bottomrule
\end{tabular}
\caption{\textbf{Update Block Weight sharing.} Learning separate update blocks---one per resolution---consistently improves the results compared to weight sharing across all update blocks.}%
\label{tab:ablation_ub_type}
\end{table}

In \cref{tab:ablation_ub_type} we compare sharing weights of the three blocks across the three resolutions of the decoder to learning separate update blocks.
As different scales exploit features of different resolutions, learning separate update blocks improves the results over weight-sharing.

\paragraph{Update Block Convolution.}
\begin{table}[t]
\centering
\begin{tabular}{clcccc}
\toprule
\multirow{2}{*}{Method} & \multicolumn{2}{c}{Sintel Clean} & \multicolumn{2}{c}{Dynamic Replica}  \\
\cmidrule(lr){2-3}\cmidrule(lr){4-5}
& \;$\delta_{3px}$\; & \;\;TEPE\;\; & \;$\delta_{1px}$\; & \;\;TEPE\;\; & \\
\midrule
Conv2D\;\; & 6.46 & 1.05 & 7.31 & 0.140 \\
\textbf{Conv3D}\;\;  & \textbf{6.24} & \textbf{0.823} & \textbf{4.76} & \textbf{0.126}  \\
\bottomrule
\end{tabular}
\caption{\textbf{Update Block Convolution.} A GRU with a 3D convolution across space and time improves the performance over the 2D variant, especially in terms of the temporal metric TEPE.}%
\label{tab:ablation_ub_conv}
\end{table}
While prior works such as CRE Stereo use 2D convolutions in the iterative update block as they operate on single time steps, we find it beneficial to extend the processing of the update block across time (\cref{tab:ablation_ub_conv}).
This results in a general improvement but shows especially large improvements in temporal consistency.

Please see the supplement for additional analysis.

\section{Conclusion}
In this paper, we make two main contributions. 
We introduce a new stereo video dataset---\emph{Dynamic Replica}---that allows training temporally consistent disparity estimators. 
Additionally, we introduce a new method that is able to improve over the state of the art in temporally consistent stereo estimation using the new dataset.

We show that other methods benefit from training on this new dataset as it contains realistic scenes and can thus reduce the domain gap between the real world and typical synthetic, abstract training datasets such as SceneFlow. 
Our combines spatial, temporal, and stereo information, enabling precise and consistent predictions across time.
In extensive ablation studies, we show that each component of the model contributes to the final performance.

\section*{Acknowledgements}
\noindent C.\ R.\ is supported by VisualAI EP/T028572/1. 
We would like to thank Filippos Kokkinos for helping to capture test videos, Roman Shapovalov and Luke Melas-Kyriazi for insightful discussions.

{\small \bibliographystyle{plain} \bibliography{main}}

\clearpage
\appendix
\section{Additional Ablations}
We ablate the proposed SST block and the choice of attention in the update block. We train the model on SceneFlow \cite{Mayer_2016} for 50k iterations with the same hyper-parameters as in the main paper.

\paragraph{SST Block Attention}
We evaluate the choice of attention types of the SST-Block in \cref{tab:ablation_sst}.
We find that including attention layers generally improves disparity estimation both in terms of accuracy and temporal consistency. 
Attention across space, stereo pairs, and time achieves the best results.
Interestingly, time attention also improves accuracy, potentially through the use of multiple viewpoints over time improving the precise location of correspondences.

\paragraph{Update Block Attention}
In \cref{tab:ablation_ub_attention} we compare different choices of attention inside the Update Block. The model with a combination of space and time attention performs well on both datasets.
Similarly, improvements are gained in both stereo and temporal metrics.

\section{Implementation details}
\noindent Here we provide additional implementation details.

\paragraph{Training}
For all the DR \& SF dataset generalization experiments, we sample the same number of \textit{frames} from both DR and SF. For temporal consistency experiments, we sample the same number of \textit{sequences} from DR and SF.

We found that learnable positional encoding for time can generalize better during inference on longer sequences. We thus use learnable encoding for time and $sin$ / $cos$ Fourier features for space.

\paragraph{Augmentations}
During training, we set image saturation to a value sampled uniformly between $0$ and $1.4$. We stretch right frames to simulate imperfect rectification: it is stretched by a factor sampled uniformly from $[2^{-0.2},2^{0.4}]$.
Following \cite{RaftTeed021}, we simulate occlusions by randomly erasing rectangular regions from each frame with probability $0.5$.

\paragraph{Inference}
For better temporal consistency during inference, we split the input video into $20$-frame chunks with an overlap of $10$ frames. We then apply the model to each chunk and discard the first and the last $5$ frames of each prediction to compose the final sequence of disparity estimations.

\paragraph{Space-Stereo-Time attention}
We add time and position encoding to left and right input feature tensors and reshape them to $(B*T, \frac{H}{16} * \frac{W}{16}, d)$. Then we apply linear self-attention~\cite{sun2021loftr} across space to both tensors and cross attention across space between left and right tensors. Finally, we reshape left and right tensors to $(B* \frac{H}{16} * \frac{W}{16}, T, d)$ and apply standard attention across time.

\paragraph{3D CNN-based GRU}
For efficiency, each 3D GRU module is composed of three separable height-width-time GRUs with kernel sizes $(1\times 1 \times 5)$, $(5 \times 1 \times 1)$, and $(1 \times 5 \times 5)$.

\paragraph{Upsampling}
To pass the output of each update block $g$ to a higher resolution update block, we use a combination of convex upsampling from RAFT \cite{RaftTeed021} and standard bi-linear upsampling.

\section{Limitations}
While our method is more temporally consistent than previous work, it still is not fully stable over time. This partially comes form the fact that the method is evaluated in a sliding window fashion resulting in low-frequency oscillations at the scale of the window size (1-2 sec). Extending the window size is currently not possible due to memory limitations.

As with any stereo-matching method, exceedingly large untextured scene parts such as walls and other surfaces are difficult to predict accurately. Learning from DynamicReplica helps to learn priors to mitigate this issue but does not solve it completely.

As dense groundtruth information is near impossible to collect, evaluation and training rely on synthetic datasets such as DynamicReplica. Generalization to the real world can only be assessed qualitatively and might not fully reflect the performance on artificial scenes.

\begin{table}[t]
\footnotesize
\centering
\begin{tabular}{ clcccc}
\toprule
\multirow{2}{*}{Attention} & \multicolumn{2}{c}{Sintel Clean} & \multicolumn{2}{c}{Dynamic Replica} \\
& Bad3px & TEPE & Bad1px & TEPE & \\
\midrule
None  &  6.47 & 0.779 & 7.31 & 0.119 \\
Space + Stereo  & 6.14  & 0.814 & 7.37 & \textbf{0.116} \\
\textbf{Space + Stereo + Time} &  \textbf{6.02} & \textbf{0.753} & \textbf{5.50} & 0.120 \\
\bottomrule
\end{tabular}
\caption{\textbf{SST-Block Attention.} We compare no attention (none), spatial (space) and time attention. }
\label{tab:ablation_sst}
\end{table}

\begin{table}[t]
\centering
\begin{tabular}{ clcccc}
\toprule
\multirow{2}{*}{Attention} & \multicolumn{2}{c}{Sintel Clean} & \multicolumn{2}{c}{Dynamic Replica} \\
& Bad3px & TEPE & Bad1px & TEPE & \\
\midrule
None  & 6.42 & 0.940 & 7.41 &  0.117  \\
Space & \textbf{5.99} & 0.864 &  7.26 &  \textbf{0.114} \\
\textbf{Space + Time} & 6.02 & \textbf{0.753} & \textbf{5.50} & 0.120 \\
\bottomrule
\end{tabular}
\caption{\textbf{Update Block Attention.} A combination of space and time attention helps to propagate information.}%
\label{tab:ablation_ub_attention}
\end{table}
\setlength\tabcolsep{.4em}
\begin{table}[t]

\footnotesize
\centering
\begin{tabular}{ lc}
\toprule
Method & sec./frame \\
\midrule
RAFT Stereo~\cite{raft_stereo} & 0.83  \\
CODD~\cite{Holograms}  &  1.04 \\
DynamicStereo (Ours)  &  1.20 \\

\bottomrule
\end{tabular}
\caption{\textbf{Runtime analysis.} We run each method on a video of resolution 1280x720 on a GPU and report the average number of seconds it takes the method to process one frame.  }
\label{tab:ablation_sst}
\end{table}

\end{document}